%% file: main.tex
\documentclass[10pt]{article}

\usepackage{etoolbox}
\newcommand{\arxiv}[1]{\iftoggle{iclr}{}{#1}}
\newcommand{\iclr}[1]{\iftoggle{iclr}{#1}{}}
\newtoggle{iclr}
\global\togglefalse{iclr}

\PassOptionsToPackage{dvipsnames}{xcolor}

\usepackage[utf8]{inputenc}
\usepackage[T1]{fontenc}
\usepackage{url}
\usepackage{booktabs}
\usepackage{amsfonts}
\usepackage{nicefrac}
\usepackage{microtype}
\usepackage{bm}
\usepackage{dsfont}
\usepackage{wrapfig}
\usepackage{enumitem}

\usepackage{colortbl}
\usepackage{xcolor}

\usepackage{inconsolata}
\DeclareTextCommandDefault{\textasciigrave}{\UseTextSymbol{TS1}\textasciigrave}
\usepackage[scaled=.90]{helvet}
\usepackage{xspace}
\usepackage[most]{tcolorbox}
\usepackage{pifont}

\input{math_commands.tex}

\input{arxiv_style}

\definecolor{citationcolor}{RGB}{50,100,170}
\definecolor{urlcolor}{RGB}{255,102,178}
\definecolor{postcutoff}{RGB}{236,235,243}
\definecolor{humanrow}{RGB}{229,242,232}
\definecolor{schemarow}{RGB}{255,232,228}
\definecolor{opgreen}{HTML}{006E45}
\definecolor{opblue}{HTML}{1A79EF}
\definecolor{opslate}{HTML}{63778F}
\input{figures/case_macros}

\newcommand{\toolout}[1]{\par\smallskip\begingroup\footnotesize\ttfamily\raggedright\leftskip=1em\rightskip=1em plus 1fil\noindent #1\par\endgroup\smallskip}
\usepackage{graphicx}

\newcommand{\citepmonth}[2]{\textcolor{citationcolor}{(\citeauthor{#2}, #1)}}

\usepackage{etoc}

\usepackage{caption}
\usepackage{parskip}
\DeclareHookRule{begindocument}{parskip}{before}{hyperref}
\usepackage{needspace}
\newcommand{\wrapneedspace}[1]{\Needspace{#1\baselineskip}}

\title{Schema: Discovering Unknown Environments \\via Agentic Program Induction}
\author{
  \textbf{Guanning Zeng}$^{3*}$ \quad
  \textbf{Jiani Wang}$^{1}$ \quad
  \textbf{Wenjie Ma}$^{2}$ \quad
  \textbf{Shaofeng Yin}$^{1}$ \\
  \textbf{Chenyang Wang}$^{1}$ \quad
  \textbf{Shichen Liu}$^{1}$ \quad
  \textbf{Angjoo Kanazawa}$^{2}$ \quad
  \textbf{Wode Ni}$^{1}$ \\
  \textbf{Xiuyu Li}$^{1*}$ \quad
  \textbf{Andrea Zanette}$^{3*}$ \quad
  \textbf{Haiwen Feng}$^{1,2*}$ \\[2mm]
  {\small
  $^{1}$Impossible Research \quad
  $^{2}$UC Berkeley \quad
  $^{3}$Carnegie Mellon University
  } \\[2mm]
  {\small Project website: \url{https://schema-harness.github.io/}}
}

\hypersetup{
  pdftitle={Schema: Discovering Unknown Environments via Agentic Program Induction},
  pdfauthor={Guanning Zeng and Jiani Wang and Wenjie Ma and Shaofeng Yin and Chenyang Wang and Shichen Liu and Angjoo Kanazawa and Wode Ni and Xiuyu Li and Andrea Zanette and Haiwen Feng},
}

\begin{document}
\etocdepthtag.toc{maintext}

\maketitle
\renewcommand{\thefootnote}{}
\footnotetext{$^{*}$Project leads.}
\renewcommand{\thefootnote}{\arabic{footnote}}

\thispagestyle{fancy}
\fancyhead{}
\lhead{\raisebox{-0.70cm}{\includegraphics[height=0.7cm]{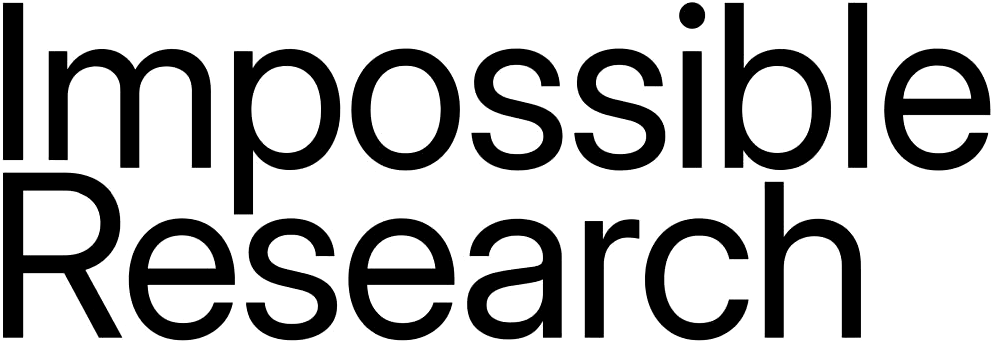}}\hspace{0.6cm}\raisebox{-0.66cm}{\includegraphics[height=0.62cm]{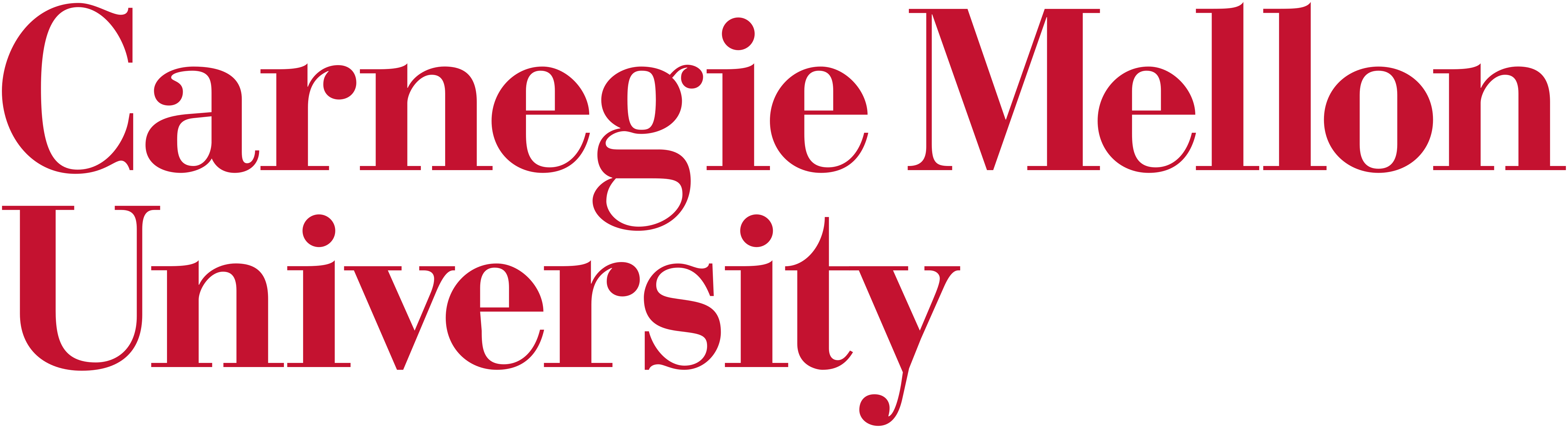}}\hspace{0.6cm}\raisebox{-0.66cm}{\includegraphics[height=0.62cm]{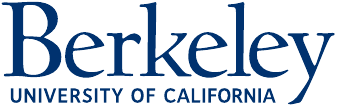}}}
\renewcommand{\headrulewidth}{0pt}

\begin{abstract}
Learning to complete tasks in unfamiliar environments with unknown rules remains a key challenge for LLM agents.
Current LLM agents often record their discoveries in prose, which may not provide a compact, explicit account of how the environment works.
Inspired by how scientists organize observations into testable, predictive theories, we introduce \textbf{Schema}, an agent harness that organizes learning and action through \emph{interactive program induction}.
The LLM agent decides what to investigate and how to act, expressing its evolving understanding of the environment as executable programs.
The harness consists of a persistent program workspace and a small set of interfaces for checking these programs against the interaction history, planning within them, and executing plans under step-by-step verification.
Schema raises ARC-AGI-3 RHAE from 58.7\% to 99.2\% with the same base model, solves 100\% of the public DiG-bench games, and reaches the median performance of the top-50 human players on MazeBench.
Extensive analysis shows the effectiveness of Schema in unknown mechanism discovery, and ablations confirm the contribution of each component.
\end{abstract}

\section{Introduction}
Provided with detailed financial data, an LLM agent can quickly produce a comprehensive summary.
Given trip destinations and a budget, it can easily work out a thoughtful itinerary.
Armed with established algorithms, it can even beat top human competitors in programming contests~\citep{openai2025icpc}.
Modern LLM agents, built on frontier models and equipped with advanced tools, now excel at tasks with familiar settings, known procedures, and well-defined objectives~\citep{jimenez2024swebench,merrill2026terminalbench,xie2024osworld,yao2024taubench,elkishky2025competitive,chan2025mlebench,paglieri2025balrog,kwa2025measuring}.
Yet acting in unknown environments where no explicit rules or instructions are given remains a challenge~\citep{chen2026testtime,chen2026why,phan2025textquests}.
For example, on benchmarks that require LLM agents to learn a completely novel environment from scratch, even the strongest models still fall short of human performance~\citep{warrier2025worldtest,battleday2026dig,pappas2026mazebench}. 

\begin{figure}[!t]
\centering
\includegraphics[width=\linewidth]{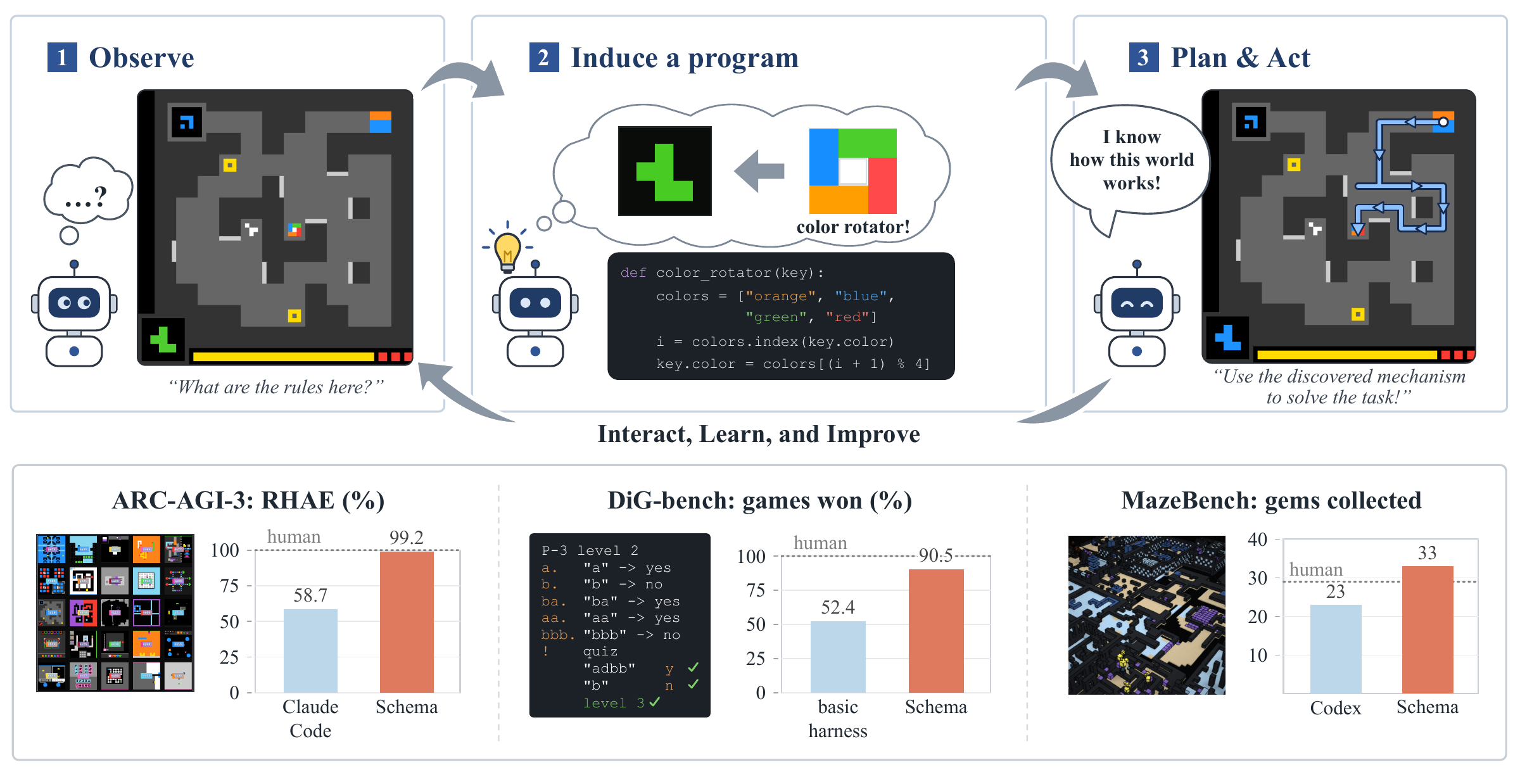}
\vspace{-4mm}
\caption{Schema observes the environment, encodes discovered mechanisms as executable programs, and uses them to guide planning and action, refining its understanding through further interaction (top). Schema achieves strong performance across all three benchmarks, outperforming baseline harnesses using the same base models (bottom); dashed lines indicate human reference performance. (a) ARC-AGI-3: RHAE over the 25 public games, Claude Fable 5 in Claude Code and in Schema; the human baseline is 100 by construction. (b) DiG-bench: public games won out of 21, GPT-6 Astra. (c) MazeBench: gems collected, GPT-6 Astra in Codex and in Schema, against the top-50 human median.
}
\label{fig:teaser}
\end{figure}

This gap reflects a difference in how LLM agents and humans discover the mechanisms of an unknown environment. 
Humans do so through observation, experimentation, and reasoning~\citep{gopnik2012reconstructing,cook2011where}, and abstract what they find into explicit, reusable theories~\citep{tenenbaum2011grow,rule2020child}. 
In scientific discovery, for example, researchers derive predictions from theories, test them through experiments, and revise the theories in light of new evidence. 
Such theories transfer across situations~\citep{lake2017building}, letting humans act efficiently and adapt quickly to new challenges~\citep{tsividis2021human,dubey2018investigating}. 
LLM agents, by contrast, hold what they learn as prose in a context window or in editable memory. 
As context is compacted and memory rewritten~\citep{packer2023memgpt}, that prose degrades~\citep{liu2024lost}, and an agent that does not distill it into a persistent, reusable form is left disoriented in a complex, new environment~\citep{laban2026lost,backlund2025vendingbench}. 

Inspired by this observation, we present \textbf{Schema}, the first agentic harness that achieves \textbf{human-level} mechanism discovery in unknown environments. 
Schema expresses its understanding of environment mechanisms in the formal language of computers: programs.
To organize how an agent constructs, tests, refines, and uses this understanding, we propose \textbf{interactive program induction}, a paradigm for agent design that Schema instantiates.
Concretely, the LLM agent writes programs that encode environment representations and transitions, checks them against its interaction history, and uses them to guide experimentation and planning (Figure~\ref{fig:teaser}).
This approach allows Schema to keep and continually update what it learns in a form that is persistent, executable, and verifiable.

We evaluate Schema's generality across environments and base models, efficiency in discovering hidden mechanisms, and ability to sustain progress over long horizons.
On ARC-AGI-3~\citep{arcprize2026arcagi3}, Schema improves action efficiency across four model configurations, reaching 99.2\% RHAE compared with 58.7\% for the same base model in its coding harness.
On DiG-bench~\citep{battleday2026dig}, Schema discovers hidden rules with fewer failed attempts and solves all 21 public games.
On MazeBench~\citep{pappas2026mazebench}, Schema accumulates and reuses knowledge across tens of thousands of interactions, sustaining exploration and reaching the median performance of the top-50 human players. Our contributions are as follows:

\begin{itemize}
   \item \textbf{Paradigm.} We propose interactive program induction, a paradigm for designing LLM agents that learn and act in unknown environments by continually constructing, testing, and using executable programs.
   \item \textbf{Harness.} We build Schema, which instantiates this paradigm and reaches human-level performance on various novel environments.
   \item \textbf{Analysis.} Through ablations and behavioral analyses, we show how Schema's components support mechanism discovery and sustained exploration.
\end{itemize}

\section{Preliminaries}
\label{sec:preliminaries}

\paragraph{Task Setting.}
\label{sec:problem}
Consider an agent acting in an unfamiliar environment whose available actions are given, but whose rules and completion conditions must be discovered through interaction.
At step $t$, the agent receives an observation $o_t$ of latent environment state $s_t$ and chooses an action $a_t\in\mathcal{A}$.
The environment evolves according to $s_{t+1}=T(s_t,a_t)$ and returns $o_{t+1}$, including task feedback.
The unknown transition function $T$ and completion predicate $G$, with $G(s)=1$ indicating success, together specify the environment's \emph{mechanisms}.
The agent accumulates the ordered interaction history $\mathcal{H}_t=((o_i,a_i,o_{i+1}))_{i<t}$.
Its objective is to complete the task using as few interactions with the environment as possible (an objective of broad practical relevance, whether imposed by game rules or motivated by the cost of real-world experimentation).
Actions therefore serve both to advance the task and to provide evidence about the environment's mechanisms.

\paragraph{Program Induction.}
\label{sec:background}
Program induction seeks an executable program $P$ consistent with a set of input--output examples $\mathcal{D}=\{(x_i,y_i)\}$, so that $P(x_i)=y_i$ for each example.
For example, from the pairs $\texttt{abc}\to\texttt{cba}$ and $\texttt{cab}\to\texttt{bac}$, a learner may infer a program that reverses its input string.
Standard counterexample-guided inductive synthesis organizes this search as a dialogue between a synthesizer and a verifier~\citep{solarlezama2006cegis,solarlezama2008thesis}.
The synthesizer proposes a candidate consistent with the current examples. The verifier checks it against a specification and returns a counterexample when it fails.
Adding that counterexample to $\mathcal{D}$ constrains the next candidate while retaining the evidence already collected.
For environment learning, action outcomes provide examples from which language models can induce programs describing the underlying dynamics~\citep{tang2024worldcoder,dainese2024code}.
Executing these programs makes their predictions available for both checking and simulation.
The choice of which actions to take thus becomes part of the induction process.

\section{Schema: Agentic Harness with Interactive Program Induction}
\label{sec:schema}
\label{sec:ipi}

\begin{figure}[t]
\centering
\includegraphics[width=\linewidth]{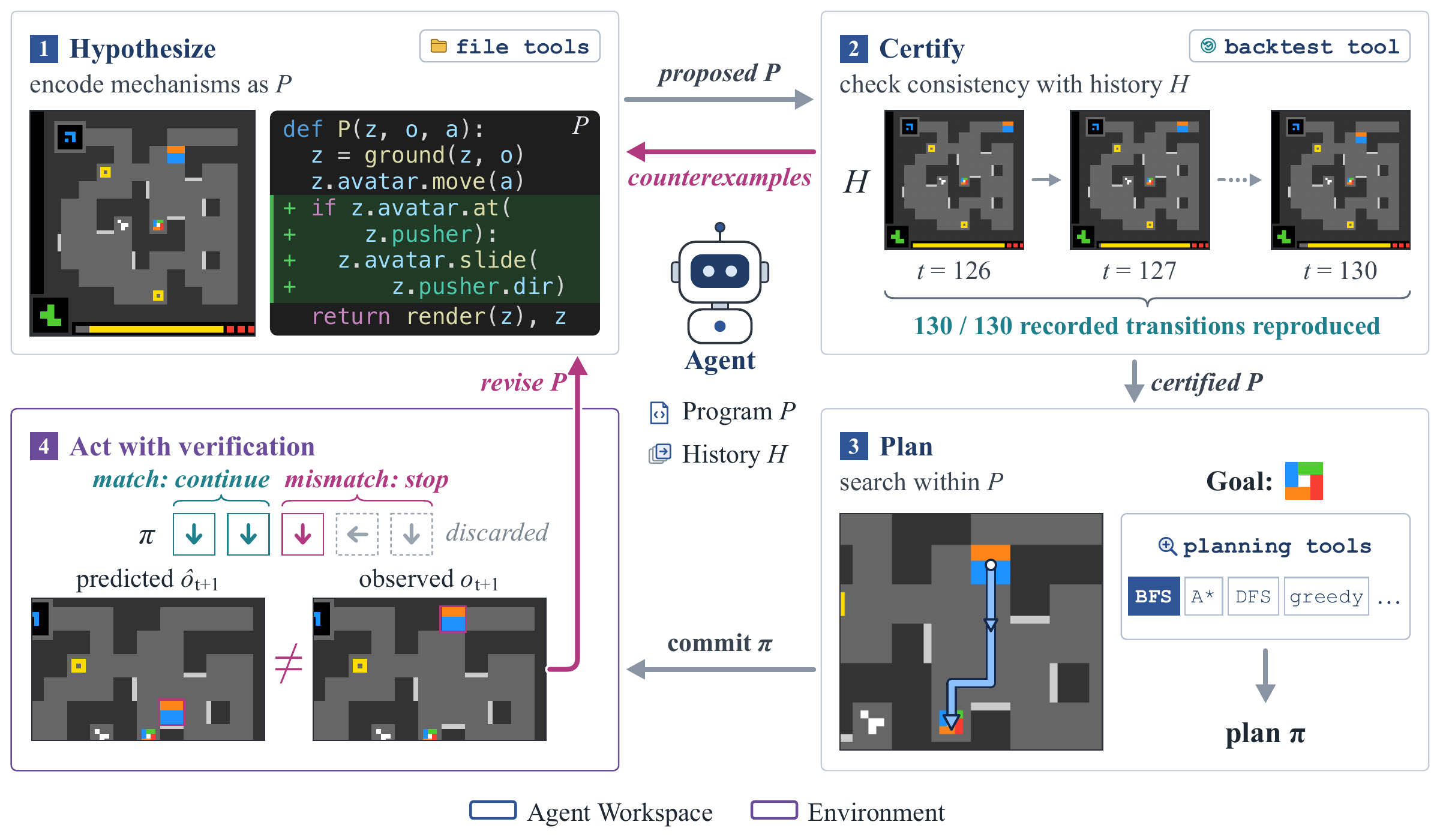}
\vspace{-3mm}
\caption{\textbf{Overview of Schema.} The LLM agent expresses its understanding as an executable program $P$ in a persistent workspace. The harness checks $P$ against recorded interactions and searches within it using goals and procedures chosen by the agent, without consuming environment actions. Committed plans execute one action at a time under prediction checks. A mismatch interrupts execution and returns new evidence for revising $P$. The ARC-AGI-3 LS20 level 3 example illustrates an unmodeled color-rotator effect with a simplified prediction: the key is predicted to remain orange but turns blue. 
}
\label{fig:schema}
\end{figure}

We propose \emph{interactive program induction} (IPI), a design paradigm that integrates program induction and action selection into a single LLM agent.
Schema instantiates IPI through the four basic operations of \emph{hypothesize}, \emph{certify}, \emph{plan}, and \emph{act with verification} (Figure~\ref{fig:schema}).
Schema's harness consists of a persistent workspace and a set of interfaces for carrying out these operations (Appendix~\ref{app:harness}).
The LLM is free to organize calls to these interfaces, drawing on its broad prior knowledge to make context-dependent decisions that guide its interaction with the environment.

\subsection{Hypothesize: Encoding Mechanisms as Programs}
\label{sec:programs}

The Schema agent encodes its current understanding of the environment in a program $P$, which typically includes two components.
(i) \emph{State grounding} constructs a structured representation of observed objects, attributes, and relations, together with relevant information carried over from earlier interactions.
(ii) \emph{Transition rules} encode the effects of actions and object interactions on this representation.
The harness invokes $P$ through a common prediction interface
\begin{equation}
\label{eq:program}
    (\hat{o}_{t+1},z_{t+1})=P(z_t,o_t,a_t),
\end{equation}
where $z_t$ is the program's internal state and $\hat{o}_{t+1}$ is its prediction of the next observation.
The agent uses file management tools to revise the program's state representation and transition rules, while the harness appends each observed transition to the interaction history.

\subsection{Certify: Checking Consistency with History}
\label{sec:certification}

A proposed program is checked by replaying the interaction history.
For each recorded action, the harness evaluates Equation~\ref{eq:program} using the recorded observation and the internal state reconstructed under $P$.
Let $\hat{o}_{i+1}^{P}$ denote this prediction, and write $\hat{o}_{i+1}^{P}\simeq o_{i+1}$ when its observable consequences agree with the environment's output, including completion or failure feedback.
The mismatch set and history-consistency criterion are
\begin{equation}
\label{eq:cert}
\begin{aligned}
    \mathcal{M}(P,\mathcal{H}_t)&=\{\,i<t:\hat{o}_{i+1}^{P}\not\simeq o_{i+1}\,\},\\
    \mathrm{Cert}(P,\mathcal{H}_t)&\Longleftrightarrow\mathcal{M}(P,\mathcal{H}_t)=\emptyset.
\end{aligned}
\end{equation}
In Schema, the harness provides backtesting tools that report mismatches together with predicted and observed outcomes, giving the agent concrete counterexamples from which to revise its hypothesis.
Rechecking the full history tests each revision against previously explained behavior, allowing discoveries to accumulate in a coherent program.
In practice, the LLM iteratively uses file management and backtesting tools to revise the program and check its consistency with the full interaction history.

\subsection{Plan: Searching within the Program}
\label{sec:loop}
\label{sec:schema-plan}

The same prediction interface makes $P$ a simulator: feeding predicted observations and internal states back into Equation~\ref{eq:program} produces a rollout without interacting with the environment.
The agent specifies a target predicate $g$ over these simulated states and observations.
The target may express task completion, an intermediate waypoint, or a situation in which an unresolved mechanism can be tested.
Planning therefore serves both task progress and experiment design.
Given a target, the agent chooses a search procedure $\sigma$ and computational budget $b$ and invokes
\begin{equation}
\label{eq:plan}
    \pi\leftarrow\textsc{Search}_{\sigma}(P,z_t,o_t,g; b),
    \qquad \pi=(a_t,\ldots,a_{t+k-1}).
\end{equation}
The Schema harness runs the selected procedure inside $P$ and returns a candidate plan or reports that none was found within the budget.
The agent can call built-in planning tools (e.g., breadth-first, depth-first, A$^*$, or greedy search) or write and load any search procedure of its own.
It also sets and can adjust search depth, node limits, and runtime in its tool calls.
The returned plan and search feedback help the agent decide whether to commit to the actions, refine the search, or investigate further.

\subsection{Act with Verification: Testing Predictions in the Environment}
\label{sec:execution}

The agent commits an action sequence of its chosen length, obtained through search or designed directly as an experiment.
Before each action $a_t$, the harness obtains the program's prediction using Equation~\ref{eq:program}; it then executes the action and appends the observed transition to $\mathcal{H}_t$.
If prediction and observation agree, execution proceeds without another LLM call.
Inspired by model predictive control~\citep{li2014eventtriggered}, the harness discards the remaining actions at the first prediction mismatch and returns the unexpected transition to the agent.
Each executed action extends the evidence available for induction.
The agent is free to use the returned observations to decide whether to revise the program, adjust its plans, or investigate further.

\section{Experiments}
\label{sec:experiments}

Our experiments assess Schema's \textit{generality across games and base models}, \textit{efficiency in discovering hidden mechanisms}, and \textit{ability to sustain progress over long horizons}.
ARC-AGI-3 establishes performance across base models and uses component ablations to explain how discoveries translate into efficient action (Section~\ref{sec:arcagi3}).
DiG-bench examines the cost of identifying hidden rules, measured by games won and lives lost at different reasoning efforts (Section~\ref{sec:digbench}).
MazeBench examines whether knowledge remains compact and useful as interaction history grows, linking program reuse to continued exploration (Section~\ref{sec:mazebench}).
Within each benchmark, we compare Schema with baseline harnesses using the same base model.
Further benchmark background and scoring details are given in Appendix~\ref{app:leaderboards}, and gameplay videos are in the supplementary material.

\subsection{ARC-AGI-3: Human-level Action Efficiency}
\label{sec:arcagi3}

ARC-AGI-3 requires agents to discover unfamiliar game mechanics through interaction and apply them to increasingly complex levels.
We assess Schema's performance relative to human players and baseline harnesses, then use component ablations to analyze the source of its gains.

\paragraph{Task Setup.}
ARC-AGI-3~\citepmonth{March 2026}{arcprize2026arcagi3} contains 25 public games, each with six to ten levels presented as 64$\times$64 color grids.
The agent is given up to seven available actions and must discover their effects and the game's objective through interaction.
The primary metric, relative human action efficiency (RHAE), combines level completion with action efficiency relative to first-time human players (Appendix~\ref{app:arc-background}).
We also report the number of games won and games reaching the maximum score.
The benchmark remains challenging: in its official basic harness, models whose training-data cutoffs precede its public release still only achieve at most 30\% RHAE.
We evaluate Claude Opus 4.8~\citep{anthropic2026opus48}, Claude Fable 5~\citep{anthropic2026fable5}, and GPT-5.6 Sol~\citep{openai2026gpt56}, with Sol tested at both extra-high and maximum reasoning effort.
We compare Schema with two types of baseline harnesses using the same base model: (i) the official ARC-AGI-3 basic harness, which provides no auxiliary tools, and (ii) the model's coding harness (Claude Code~\citep{anthropic2026claudecode} for Claude models and Codex~\citep{openai2026codexcli} for GPT models), which allows agents to maintain persistent notes and execute code and shell commands.

\begin{figure}[t]
\centering
\includegraphics[width=\linewidth]{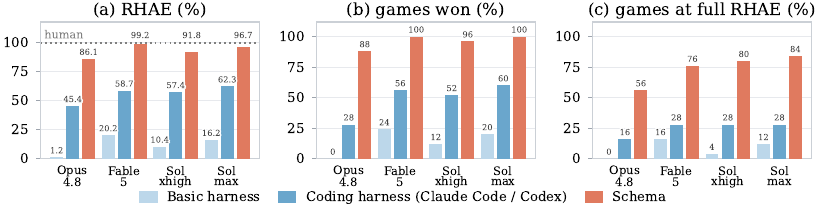}
\vspace{-3mm}
\caption{\textbf{Schema improves completion and action efficiency across model configurations.} Results on the 25 public ARC-AGI-3 games: (a) RHAE, (b) percentage of games won, and (c) percentage of games achieving 100\% RHAE. Each model configuration is evaluated with the basic harness, coding harness, and Schema.}
\label{fig:arc}
\end{figure}

\wrapneedspace{12}
\begin{wraptable}{r}{0.30\textwidth}
\vspace{-2\baselineskip}
\centering
\small
\setlength{\tabcolsep}{4pt}
\setlength{\belowcaptionskip}{6pt}
\caption{Ablation results.}
\label{tab:ablation}
\begin{tabular}{lr}
\toprule
& RHAE (\%) \\
\midrule
\rowcolor{schemarow}
\textbf{Schema (full)} & \textbf{72.9} \\
prose model       & 58.8 \\
w/o certification & 62.3 \\
w/o planning      & 59.2 \\
w/o verification  & 52.4 \\
\bottomrule
\end{tabular}
\end{wraptable}
\paragraph{Main Results.}
Across the four configurations, Schema raises RHAE from 1.2--20.2\% in the basic harness to 86.1--99.2\% and increases games won from 0--6 to 22--25 (Figure~\ref{fig:arc}).
Relative to the coding harnesses, Schema improves RHAE by 34.4--40.7 percentage points (Figure~\ref{fig:arc}a).
It reaches 99.2\% with Fable 5, 96.7\% and 91.8\% with Sol at maximum and extra-high effort, and 86.1\% with Opus 4.8, compared with 58.7\%, 62.3\%, 57.4\%, and 45.4\% in their coding harnesses.
With Fable 5 and Sol at maximum effort, Schema clears all 25 games, compared with 14 and 15 for the respective coding harnesses; 19 and 21 games reach the maximum score, compared with seven for each baseline (Figure~\ref{fig:arc}b,c). With Fable 5, Schema completes every one of the 25 games in fewer actions than the human baseline, even using just 57\% of the human action total across all 183 levels.
These results suggest that frontier models do not lack the ability to understand unfamiliar environments. However, they need the right harness to put their broad knowledge to use.

\setlength{\columnsep}{18pt}%
\paragraph{Ablation Study \& Analysis.}
To further understand how the harness helps models put their knowledge to use, we replace the executable program with prose and separately remove the interfaces for certification, planning, and verified execution from Schema.
We evaluate all variants with Claude Opus 4.8 on the ten hardest games, ranked by the number of actions in their human baselines, covering 75 levels with a budget of 2,000 actions per game.
As shown in Table~\ref{tab:ablation}, removing any of these components substantially reduces performance, with RHAE falling from 72.9\% to 52.4--62.3\%.
\setlength{\columnsep}{10pt}%

\begin{figure}[t]
\centering
\includegraphics[width=\linewidth]{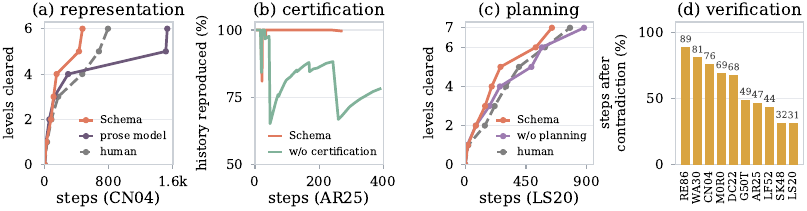}
\vspace{-3mm}
\caption{\textbf{Trace dynamics.} (a) Levels cleared versus cumulative steps on CN04 with program and prose representations, alongside humans. (b) Offline history agreement on AR25 with and without certification, using Schema's Opus 4.8 main-evaluation run. (c) Levels cleared versus cumulative steps on LS20 with and without planning, alongside humans. (d) Steps taken after a prediction mismatch within the same plan without verification.}
\label{fig:arc-ablation}
\end{figure}

We further examine the agents' reasoning and interaction traces to understand how these ablations affect their discovery and use of the environment's rules, with detailed cases in Appendix~\ref{app:case-arc-analysis}.
(i) \textbf{Program induction makes beliefs explicit and reusable.}
In CN04 level 5, pieces must be moved and rotated to pair their connectors without overlapping, while one piece grows and changes the geometry.
The prose agent repeatedly revisits the connection rules; Schema enumerates arrangements under explicit geometric constraints, completing the level in 279 rather than 1,219 actions (Figure~\ref{fig:arc-ablation}a).
(ii) \textbf{Certification helps the agent validate hypothesis correctness.}
In AR25, the agent moves pieces and mirror axes so that reflected shapes match a target.
Without certification, a revision to identify the active piece improves predictions on the current level but drops history agreement from 97.7\% to 65.2\%; Schema's final model reproduces 99.6\% of its history (Figure~\ref{fig:arc-ablation}b).
(iii) \textbf{Planning ahead improves action efficiency.} Even with a program that fits observed transitions, choosing an efficient action sequence requires reasoning about future states.
In LS20 level 5, the agent must transform a carried symbol to match its target, timing contact with a moving rotation marker within an energy budget.
Without planning, the model passes all 252 transition checks at entry to the level, but the agent works out routes and transformations manually, taking 233 actions compared with 65 for Schema (Figure~\ref{fig:arc-ablation}c).
(iv) \textbf{Runtime verification further boosts adaptation to new scenarios.}
In LS20 level 2, the agent must rotate a carried symbol and reach the goal using single-use refueling stations.
Without verification, it assumes these stations are reusable and continues for 42 actions after the first mismatch, returning to a spent station and running out of energy.
Across the ten games in Figure~\ref{fig:arc-ablation}d, such continuations account for 58\% of actions.

\subsection{DiG-bench: Efficient Mechanism Discovery}
\label{sec:digbench}

We next turn to a setting where discovering hidden mechanisms is central to success.
On DiG-bench, we examine whether Schema can choose informative interactions and efficiently infer the underlying rules.
\paragraph{Task Setup.}
DiG-bench~\citepmonth{August 2026}{battleday2026dig} evaluates interactive mechanism discovery through text-based games across seven difficulty tiers.
Game states are presented as short strings of letters, numbers, and symbols, but both the transition rules and the win conditions must be discovered through interaction.
For example, in P-3, the agent queries whether chosen strings satisfy a hidden rule, then classifies new strings, with each incorrect quiz answer costing a life.
The agent must choose informative actions and use the resulting feedback to identify hidden rules while completing challenges with very limited lives and per-level step budgets.
Some games also offer a creative mode for experimentation, allowing the agent to construct situations that test its hypotheses.
These games remain challenging for frontier models.
The benchmark authors report that the strongest model achieves a 71.4\% win rate overall and only 40\% on the two hardest tiers, while every game was solved by at least one human on their first attempt.
Existing agentic harnesses have also shown limited gains over the basic harness on these discovery tasks~\citep{battleday2026dig}.
We evaluate Schema on the 21 publicly available games, spanning all seven difficulty tiers, using GPT-6 Astra~\citep{openai2026gpt6} at medium, high, and maximum reasoning effort.

\begin{figure}[t]
\centering
\includegraphics[width=\linewidth]{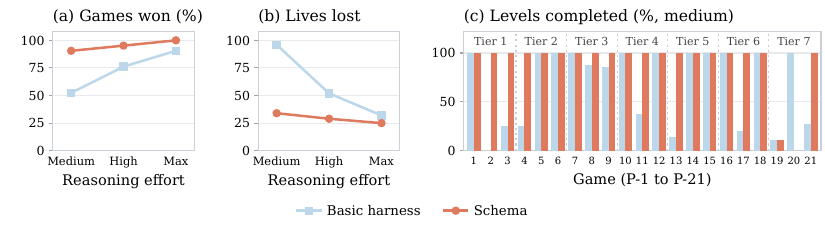}
\vspace{-3mm}
\caption{\textbf{Schema solves more games with fewer failed attempts.} GPT-6 Astra on the 21 public DiG-bench games: (a) game win rate and (b) total lives lost at each reasoning effort; (c) percentage of levels completed in each game at medium effort, grouped by difficulty tier.}
\label{fig:dig}
\end{figure}

\begingroup
\setlength{\intextsep}{0pt}
\wrapneedspace{16}
\begin{wrapfigure}{r}{0.50\textwidth}
\centering
\includegraphics[width=\linewidth]{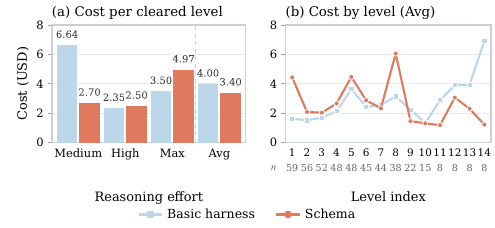}
\caption{\textbf{Token cost on DiG-bench.} Average cost (a) per cleared level and (b) by level index, including input and output tokens at OpenAI's official API rates~\citep{openai2026gpt6pricing}.}
\label{fig:dig-cost}
\end{wrapfigure}
\paragraph{Results.}
Compared with the basic harness, Schema improves game completion and reduces failed attempts at all three reasoning efforts (Figure~\ref{fig:dig}).
It clears 19, 20, and 21 games at medium, high, and maximum effort, respectively, compared with 11, 16, and 19 for the basic harness.
On the two hardest tiers (6--7), Schema raises the win rate at maximum effort from 66.7\% to 100\%.
At medium effort, Schema already matches the basic harness's maximum-effort win count; compared with the basic harness at medium effort, it also reduces lives lost from 96 to 34.
The same trends hold across repeated runs (Appendix~\ref{app:dig-repeats}).
Pooling costs and completed levels across all three reasoning efforts, Schema's estimated token cost per cleared level is \$3.40, approximately 15\% lower than the basic harness's \$4.00 (Figure~\ref{fig:dig-cost}a).
For levels cleared by both harnesses, the cost curve pooled by level index shows higher spending by Schema at early levels but lower mean costs at every index from 9 to 14 (Figure~\ref{fig:dig-cost}b).
This pattern is consistent with reusing mechanisms discovered early in the game to solve later levels.
Together, these results show that Schema supports efficient mechanism discovery with fewer failed attempts, a lower aggregate token cost per cleared level, and less reasoning effort to reach comparable game completion.
\par\endgroup

\paragraph{Analysis.}
We examine the interaction traces and find that Schema is more systematic than the basic harness in testing its hypotheses.
When feedback contradicts its current understanding, Schema revises the program and uses further queries to distinguish competing explanations, checking that the new rule also accounts for earlier observations.
This helps it resolve uncertainty before risking lives on an untested answer.
For example, in P-3 level 3, Schema tests strings with different letter orders and counts to discover the acceptance rule that the basic harness misses.
See Appendix~\ref{app:case-dig} for the queries, program revisions, and quiz outcomes in this level.

\subsection{MazeBench: Knowledge Accumulation over Long Horizons}
\label{sec:mazebench}

Finally, we examine whether Schema can build up and draw on its understanding of a large environment over long horizons. We observe sustained knowledge acquisition and agentic exploration.

\paragraph{Task Setup.}
MazeBench~\citepmonth{July 2026}{pappas2026mazebench} is a three-dimensional exploration and puzzle-solving environment with 256 interconnected rooms and 100 gems (Figure~\ref{fig:maze-env}).
Its rooms feature mechanisms such as lifts, bridges, springs, and carpets, alongside scenery ranging from forests and icy mountains to palaces and city walls; the rules governing these mechanisms must be discovered through interaction.
We evaluate Schema with GPT-6 Astra, measuring exploration progress by gems collected and rooms visited, and compare its trajectory with that of the same model in Codex, using human performance as a reference.
At submission, the top-50 human players on the official leaderboard have an average recorded playtime of 37.2 hours and an average action count of 65,876; their median gem collection and room coverage are 29.0\% and 53.5\%, respectively, reflecting the demands of sustained exploration in this environment.

\begingroup
\setlength{\intextsep}{0pt}
\wrapneedspace{13}
\begin{wrapfigure}[11]{r}{0.43\textwidth}
\centering
\setlength{\abovecaptionskip}{4pt}
\vspace{4pt}
\includegraphics[width=\linewidth]{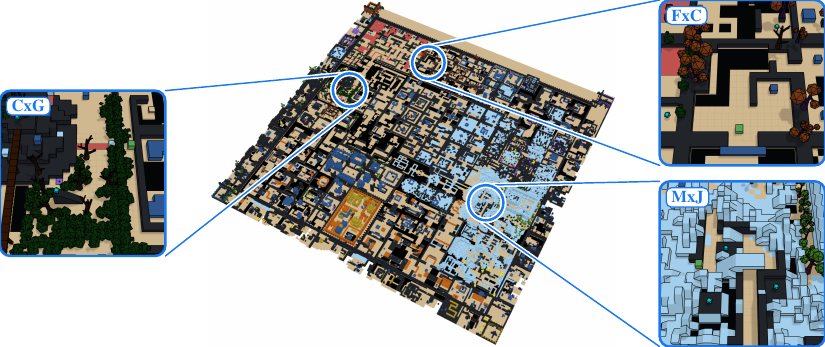}
\caption{\textbf{MazeBench.} The full world of 256 rooms, with close-ups of rooms CxG, FxC, and MxJ.}
\label{fig:maze-env}
\end{wrapfigure}
\begin{figure}[t]
\centering
\includegraphics[width=\linewidth]{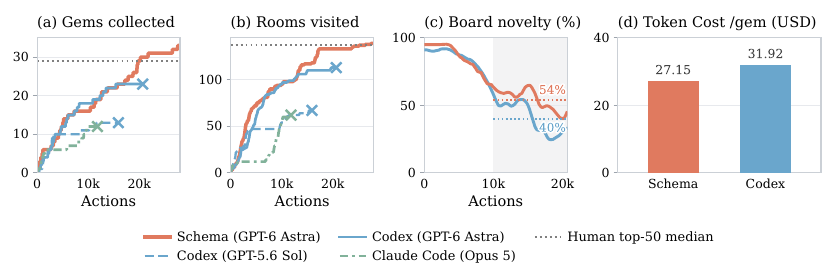}
\vspace{-3mm}
\caption{\textbf{Schema sustains progress over long horizons.} (a, b) Gems collected and rooms visited over environment actions; crosses mark the last recorded action of each baseline trajectory, and dotted lines show the top-50 human median. (c) Board-state novelty, with horizontal lines indicating each curve's mean over the shaded interval after 10,000 actions. (d) Token cost per gem at 23 collected gems.}
\label{fig:maze}
\end{figure}
\paragraph{Results.}
Over a 36-hour run, Schema collects 33 gems and visits 139 rooms in 27,819 actions, exceeding the top-50 human median on both metrics (Figure~\ref{fig:maze}a,b).
At the same action count of 20,673, Schema has collected 30 gems and visited 133 rooms, compared with 23 gems and 113 rooms for the Codex baseline, at a similar token cost (Figure~\ref{fig:maze}d).
Both agents have visited 95 rooms by action 10,000; over the next 10,000 actions, Schema adds 38 rooms and 13 gems, while Codex adds 18 rooms and five gems.
In the plotted interval after action 10,000, Schema also maintains higher mean board-state novelty, at 54\% compared with 40\% for Codex (Figure~\ref{fig:maze}c).
These results show that Schema sustains broader exploration and continues to make progress over long interaction horizons.
Appendix~\ref{app:maze-trajectory} reports every gem's location and collection step, together with room coverage maps for all four runs in Figure~\ref{fig:maze}.
\par\endgroup

\begin{figure}[t]
\centering
\includegraphics[width=\linewidth]{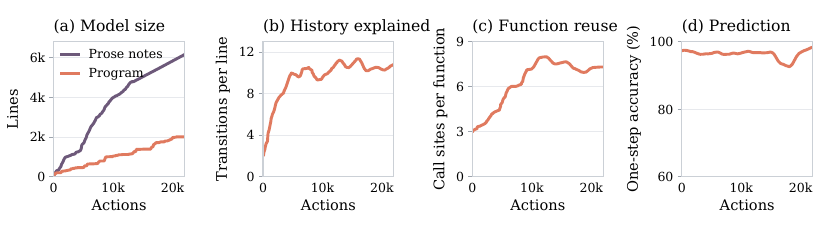}
\vspace{-3mm}
\caption{\textbf{Knowledge accumulates in a compact, reusable program.} (a) Non-comment program lines in Schema and note lines in the prose-model variant. (b) Recorded transitions explained per program line. (c) Call sites per function. (d) Forward prediction accuracy over the trailing 1,000 actions.}
\label{fig:maze-code}
\end{figure}

\paragraph{Analysis.}
We examine how the agent (i) \textit{accumulates knowledge in its program} and (ii) \textit{reuses that knowledge in new situations during prolonged interaction} with an unfamiliar environment.
Schema incorporates new experience by revising and extending the program's shared rules.
By approximately 20,000 actions, the program contains 2,012 non-comment lines, compared with 6,133 lines of notes in the prose-model variant (Figure~\ref{fig:maze-code}a).
The number of recorded transitions explained per line and the average number of call sites per function both increase, while forward prediction accuracy remains high (Figure~\ref{fig:maze-code}b--d).
These trends show that growing experience is captured in compact, reusable rules that retain their predictive value.
These rules also support planning in new configurations after repeated context compaction.
For example, on ice the player keeps sliding until blocked or reaching ordinary ground, so a route to the gem must account for where each slide stops.
Schema learns this mechanism in room IxG at around action 2,600.
Roughly 25,000 actions later, after at least 20 context compactions, it reuses the rule in a new ice maze in room NxE.
After incorporating the new room's layout into the program, Schema uses A* search to find a 38-action route to the gem.
It executes the entire route, with every observed transition matching the program's prediction (Appendix~\ref{app:case-maze}).

\section{Related Work}
\label{sec:related}

\paragraph{Agentic Harness.}
Agent harnesses shape how language models behave and what they can accomplish.
ReAct~\citep{yao2023react}, CodeAct~\citep{wang2024codeact}, and SWE-agent~\citep{yang2024sweagent} demonstrate the importance of interaction loops and tool interfaces.
Reflexion~\citep{shinn2023reflexion} and ACE~\citep{zhang2025agentic} enable agents to learn from experience by turning feedback into reusable reflections and strategies.
Recent work further explores self-improving agents and harnesses by revising the agent's own code~\citep{zhang2025darwin} and updating persistent prompts, memories, and skills~\citepmonth{August 2026}{karten2026primeagent}.
Schema organizes learning and action around one persistent, evolving executable theory of the environment, continually certified against the full interaction history and directly used for experimentation and planning.

\paragraph{Code World Models.}
Programs provide interpretable, executable models of environment dynamics, an approach explored in theory-based reinforcement learning before the rise of LLM agents~\citep{tsividis2021human}.
WorldCoder~\citep{tang2024worldcoder} and CWM~\citep{dainese2024code} use LLMs and execution feedback to synthesize and refine Python world models.
Subsequent work develops hierarchical planning in TheoryCoder~\citep{ahmed2025synthesizing} and compositional probabilistic representations in PoE-World~\citep{piriyakulkij2025poe}.
These methods use LLMs as operators for generating and revising code within predefined learning procedures, such as WorldCoder's REx search, and typically learn dynamics over supplied state representations.
Schema builds on these insights but makes the entire learning process agent-directed, transferring control over model revision, experimentation, and planning to the agent.
The agent determines how to learn in environments where state grounding, transition rules, and task goals are all initially unknown.

\paragraph{Concurrent Work.}
Recent work has explored related approaches on ARC-AGI-3, including EWM~\citepmonth{May 2026}{rodionov2026executable}, OPINE-World~\citepmonth{July 2026}{courtis2026opine}, NOOA~\citepmonth{July 2026}{furgale2026nvidia}, Tycho~\citepmonth{July 2026}{lehmann2026tycho}, and Twin~\citepmonth{August 2026}{skoutnev2026twin}.
These systems combine executable modeling, interaction feedback, and planning, with some also citing earlier releases of Schema.
These concurrent efforts highlight the promise of interactive program induction, which our study explores as a general approach to agent design across diverse unfamiliar environments.

\section{Conclusion}
\label{sec:conclusion}

We introduced interactive program induction, an approach in which an LLM agent learns about an unfamiliar environment by constructing, testing, and using executable models. Schema implements this approach through a persistent program workspace and tools for checking predictions against interaction history, planning, and verifying actions, while the agent directs experimentation and program revision. Across ARC-AGI-3, DiG-bench, and MazeBench, Schema improves task performance with frozen language models, supporting efficient mechanism discovery and sustained exploration over long interaction horizons. Ablations support the contribution of each component, while behavioral analyses illustrate how learned rules can persist across context compactions and be reused in new situations.
Together, these findings suggest that persistent executable models provide a way for agents to accumulate knowledge over interaction, turning individual discoveries into explicit hypotheses that can be tested, refined, reused, and ultimately acted upon.

\section*{Acknowledgements}

This work was supported in part by the National Science Foundation under Grant CCF-2106778.
The authors thank Zhiqi Chen and Emma Chu for helpful discussion and feedback.

\clearpage
\bibliography{iclr2027_conference}

\clearpage
\appendix
\etocdepthtag.toc{appendix}
\phantomsection
\section*{Appendix Contents}
\begingroup
\small
\etocsettagdepth{maintext}{none}
\etocsettagdepth{appendix}{subsubsection}
\etocsettocstyle{}{}
\tableofcontents
\endgroup
\newpage
\section{Extended Related Work}
\label{app:related}

\subsection{Agent Harnesses and Knowledge Representation}

An agent's harness shapes both how it interacts with the world and what it carries forward from those interactions.
ReAct~\citep{yao2023react} interleaves reasoning and actions, CodeAct~\citep{wang2024codeact} uses executable code as an action interface, and SWE-agent~\citep{yang2024sweagent} designs interfaces for navigating and modifying software repositories.
Learning from these interactions also requires a representation in which useful discoveries can accumulate.
Reflexion~\citep{shinn2023reflexion} retains verbal reflections on task feedback, while ACE~\citep{zhang2025agentic} incrementally develops a playbook of strategies and lessons.
Voyager~\citep{wang2023voyager} stores executable skills in a growing library, allowing previously learned behaviors to support new tasks.
Self-improving systems extend what the agent can revise: the Darwin G\"odel Machine~\citep{zhang2025darwin} modifies its own agent code, and Prime Agent~\citepmonth{August 2026}{karten2026primeagent} updates persistent prompts, memories, and skills within a programmable harness.

The interaction record itself can also serve as a durable resource.
PRO-LONG~\citepmonth{July 2026}{fox2026prolong} preserves a complete, structured history that an agent can query with code; its agents sometimes also construct transition models and search them to plan.
VISTA~\citepmonth{August 2026}{han2026vista} combines direct visual observations, free-form language reasoning, and lossless visual memory to support interaction.
Schema makes an evolving executable theory the persistent object around which learning and action are organized.
The history supplies evidence for checking that theory, and the theory supplies predictions for experiments and plans.
This gives accumulated knowledge an operational role: a discovered rule can be tested against earlier observations and used to simulate situations the agent has not yet encountered.
The program remains available across context compactions, so later decisions can build directly on earlier discoveries.

\subsection{Code World Models}

Model-based reinforcement learning uses a model of the environment to support decision-making and learning.
Dyna~\citep{sutton1990dyna} integrates learning from real experience with updates based on model-generated transitions.
MuZero~\citep{schrittwieser2020muzero} combines a learned dynamics model with tree search, predicting rewards, values, and policies for planning, while DreamerV3~\citep{hafner2025dreamerv3} learns behavior through trajectories imagined in a neural world model.
Schema shares the use of model predictions to make interaction more productive.
It accumulates environment knowledge through revisions to an explicit program while keeping the base language model frozen, making each change available for inspection, replay, and subsequent planning.

Explicit representations of environment mechanisms also have a long history in reinforcement learning.
Symbolic approaches learn relational rules describing action effects~\citep{pasula2007learning}, and theory-based reinforcement learning induces structured theories of objects, interactions, and termination conditions for exploration and planning~\citep{tsividis2021human}.
Programs offer a representation in which such knowledge can be inspected, revised, and executed to predict future states.

LLMs make it possible to synthesize these models in general-purpose programming languages.
WorldCoder~\citep{tang2024worldcoder} learns Python transition and reward functions, using consistency with experience and an optimistic planning criterion to guide program refinement through REx search.
CWM~\citep{dainese2024code} synthesizes simulators from offline trajectories, using execution feedback within a Monte Carlo tree search over code generation, improvement, and repair.
TheoryCoder~\citep{ahmed2025synthesizing} combines symbolic high-level planning with simulation in a learned program; TheoryCoder-2~\citep{ahmed2026learning} further learns reusable abstractions from experience.
Other work expands the model class and learning setting.
PoE-World~\citep{piriyakulkij2025poe} combines programmatic experts into a probabilistic transition model, while One Life to Learn~\citep{khan2025one} studies symbolic modeling of stochastic environments from unguided exploration.
Alice~\citep{seo2026baba} uses model revisions that explain new transitions but fail on earlier ones to identify missing distinctions and guide exploration.

Schema builds on this foundation at the level of agent architecture: the LLM controls when and how to revise its theory, check it against experience, investigate an uncertain mechanism, or plan toward a goal.
This control also covers the program's state grounding, allowing the agent to develop its representation alongside its transition rules and understanding of task completion.
Concurrent work on ARC-AGI-3 likewise investigates executable modeling, interaction feedback, and planning~\citep{rodionov2026executable,courtis2026opine,furgale2026nvidia,lehmann2026tycho,skoutnev2026twin}.
Our study explores this direction as a general approach to agent design across different forms of unfamiliar environments.

\subsection{Program Induction and Active Experimentation}

Program induction searches for executable explanations of examples, often using a structured program space to make this search tractable.
Programming-by-example methods synthesize programs within domain-specific languages; for example, the string-transformation system of \citet{gulwani2011automating} infers programs from input--output examples.
Sketching~\citep{solarlezama2006cegis,solarlezama2008thesis} completes partial programs against a specification, using counterexamples from a verifier to constrain subsequent candidates.
DreamCoder~\citep{ellis2021dreamcoder} learns both a library of reusable program components and a neural search policy, allowing abstractions discovered across tasks to guide later synthesis.
These approaches develop complementary ways to construct, check, and reuse programs as representations of knowledge.

Language models provide a further source of candidate hypotheses.
Hypothesis Search~\citep{wang2024hypothesis} translates language hypotheses into executable programs and evaluates them on examples.
\citet{qiu2024phenomenal} similarly study iterative hypothesis refinement, showing how symbolic interpretation helps evaluate and apply rules proposed by language models.
Schema places this synthesis-and-checking process within ongoing interaction: recorded transitions provide tests for successive program revisions, and prediction failures supply concrete evidence for further induction.
Replaying the full history checks whether a proposed repair also preserves the mechanisms learned earlier.

In an interactive environment, the learner also chooses which evidence to collect.
\citet{piriyakulkij2024doing} combine language hypotheses with probabilistic inference and information-theoretic experiment design in a rule-discovery task.
BoxingGym~\citep{gandhi2025boxinggym} studies experimental design and model discovery together in environments defined by generative probabilistic models.
Schema connects this process to action in an unfamiliar environment: the agent can use its current program to reach a situation that tests a suspected mechanism, compare the outcome with its prediction, and incorporate the resulting evidence into the same theory used for task planning.
Experimentation and task progress therefore draw on a common, continually revised account of how the environment works.

\begingroup
\raggedbottom
\section{Implementation Details}
\label{app:harness}

Schema gives a frozen language model a persistent workspace and tools for the four operations in Section~\ref{sec:schema} (Figure~\ref{fig:schema}).
Benchmark adapters supply observations, actions, and task feedback; the agent decides which operation to invoke next.
Below, the program contract follows Equation~\ref{eq:program}, and tools are grouped by function.

\paragraph{Program Workspace.}
The workspace stores $P$, notes, helper scripts, and $\mathcal{H}_t$, which the harness extends after every executed action.
These artifacts persist across turns, levels, and context compactions.
The internal state $z_t$ represents the current situation, including information inferred from past observations.
The harness reconstructs it by replaying the relevant history under the current program, with initialization at level entries and resets as defined by the benchmark.
Learned rules therefore remain available as the agent grounds them in each new situation.

A turn supplies the current observation, available actions, progress, and remaining budgets where applicable.
The agent invokes tools as needed and ends deliberation by committing actions.
Execution returns a new observation and a report of the actions taken and any event that interrupted the sequence.

The program defines how observations map to objects, attributes, and relations, and how information needed across successive actions is represented in $z_t$.
Given $(z_t,o_t,a_t)$, the program returns $(\hat{o}_{t+1},z_{t+1})$ as in Equation~\ref{eq:program}.
The predicted observation includes task feedback such as level completion, failure, or game completion.
Recorded observations support replay checks; predicted observations support simulated rollouts.
A target predicate $g(z,o)$ specifies a desired outcome: task completion, an intermediate waypoint, or a situation in which an uncertain rule can be tested.

\paragraph{Hypothesize.}
File tools let the agent read, write, edit, search, and organize its workspace.
It can revise the state representation and transition rules, maintain candidate programs, and develop helper scripts for analyzing observations.
The \texttt{read\_history} tool retrieves recorded transitions by level, index, range, or event type, with summaries or detailed before-and-after observations.
This lets the agent revisit the evidence for a rule and examine transitions that its current program does not explain.

\paragraph{Certify.}
The \texttt{backtest} tool replays recorded interactions under the program and compares its predictions with observed outcomes.
Replay covers the recorded levels by default; a narrower scope helps localize a mismatch.
The agent can also test a candidate file before adopting it.
Feedback includes prediction accuracy, mismatched indices, and predicted versus observed outcomes.
The \texttt{model\_predict} tool evaluates a selected action from the current observation and reconstructed program state, allowing the agent to inspect a prediction before acting.
Replay checks the observations produced during gameplay and the task feedback indicating completion or failure.
For ARC-AGI-3, level changes and terminal screens are checked through their corresponding outcome flags.
In the color-rotator example of Figure~\ref{fig:schema}, backtesting identifies the unexplained transition and confirms the revised rule against the accumulated history. Abbreviated reports are:
\toolout{backtest [all transitions]: 138/139 transitions fully correct; 1 mismatch\\
\ \ mismatched transitions (index:kind): \#139:observation\\
\ \ \#139 action=2 \ recorded before / after, predicted after ...\\[2pt]
backtest [all transitions]: 139/139 transitions fully correct; 0 mismatches}

\paragraph{Plan.}
The agent selects a target and searches over rollouts of $P$ from the current state, without consuming environment actions (Equation~\ref{eq:plan}).
It can use built-in procedures such as BFS and A*, or develop planning code in the workspace.
The agent chooses the target, candidate actions, and computational budget, and can supply a heuristic or intermediate waypoint.
For ARC-AGI-3, it can select click coordinates and allow a plan to begin with a level reset.
A successful search returns an action sequence and search statistics; an unsuccessful search reports whether the explored space was exhausted or a budget was reached.
The agent uses this feedback to adjust the search, investigate its model, or commit a plan.
For the level in Figure~\ref{fig:schema}:
\toolout{BFS: goal in 15 steps via level\_up; expanded 2496 nodes, 795 distinct states\\
Plan (-> commit\_actions): [2, 2, 4, 1, 1, 1, 1, 4, 1, 4, 1, 1, 3, 3, 3]}

\paragraph{Act with verification.}
The \texttt{commit\_actions} tool ends deliberation and submits an action sequence, obtained through planning or chosen directly by the agent for exploration.
The harness executes committed actions one at a time, records each transition in $\mathcal{H}_t$, and checks the program's prediction against the outcome.
Matching predictions allow execution to continue without another LLM call; a mismatch discards the remaining actions and returns the unexpected transition to the agent.
Task events such as level completion or failure also return control to the agent.
Exploration begins with individual actions whose outcomes provide evidence for building the program.
An example interruption report is:
\toolout{world model MISPREDICTED the step just taken (action 2); the rest of the committed plan was dropped. }

ARC-AGI-3 observations are grids of hexadecimal color indices, with rendered frames available for inspection; DiG-bench supplies its native text observations; MazeBench supplies an ASCII rendering of the current room.
Each adapter exposes the benchmark's native actions and task feedback, allowing the same operations to support program revision, replay, planning, and verified execution across the three environments.

\endgroup
\begingroup
\raggedbottom
\section{Benchmark Details}
\label{app:leaderboards}
We describe the three benchmarks below and summarize their published model results in Tables~\ref{tab:arcagi3}--\ref{tab:mazebench}.

\subsection{ARC-AGI-3}
\label{app:arc-background}

ARC-AGI-3~\citep{arcprize2026arcagi3} presents visual games as 64$\times$64 grids, with mechanics and objectives that must be discovered through interaction.
Successive levels recombine these mechanics into increasingly difficult puzzles, testing whether an agent can turn its discoveries into efficient plans.
The benchmark contains 25 public games, 55 semi-private games for organizer-run evaluations, and 55 private competition games.
The latter two sets are not released for general academic use; we evaluate all 25 public games, comprising 183 levels (Figure~\ref{fig:arc-games}).
The ablations in Section~\ref{sec:arcagi3} use the ten games with the most actions in their human baselines, marked in Table~\ref{tab:arc-pergame}; together they contain 75 levels.

\begin{figure}[!ht]
\centering
\includegraphics[width=0.84\linewidth]{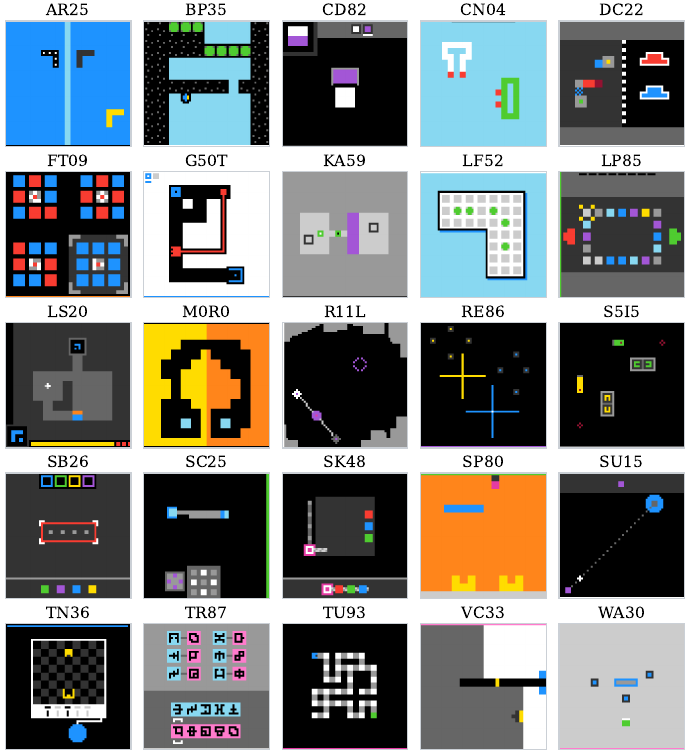}
\caption{The 25 public ARC-AGI-3 games, each shown at the start of its first level.}
\label{fig:arc-games}
\end{figure}

Relative human action efficiency (RHAE) combines level completion with action efficiency relative to first-time human play.
For a game with $L$ levels of which the agent completes the first $k$, the score is
\begin{equation}
\label{eq:rhae}
\mathrm{RHAE} = \min\!\Bigg(\frac{\sum_{l=1}^{k} l}{\sum_{l=1}^{L} l},\ \frac{\sum_{l=1}^{k} l \cdot \min\!\big(1.15,\ (h_l/a_l)^2\big)}{\sum_{l=1}^{L} l}\Bigg),
\end{equation}
where $h_l$ and $a_l$ are the human baseline's and agent's action counts on level $l$.
Later levels receive greater weight, and exceeding the human action count incurs a quadratic penalty.
We report the mean score over games as a percentage.

\begin{table}[!ht]
\centering
\footnotesize
\setlength{\tabcolsep}{4pt}
\caption{Frontier models on ARC-AGI-3, from ARC Prize's reports and leaderboard as of September 2026~\citep{arcprize2026arcagi3,arcprize2026leaderboard,arcprize2026gpt55opus47,arcprize2026solresults,arcprize2026opus5results}. Model dates follow the cited vendor documentation; empty cells are unreported. $^\dagger$Approximate public score for Fable-class models~\citep{arcprize2026fable}. $^\S$Parentheses show the provider-adapter harness~\citep{arcprize2026astra}.}
\label{tab:arcagi3}
\begin{tabular}{llccc}
\toprule
& & & \multicolumn{2}{c}{RHAE (\%)} \\
\cmidrule(lr){4-5}
Model & Cutoff & Release & Public & Semi-private \\
\midrule
Gemini 3.1 Pro~\citep{google2026gemini31pro} & Jan 2025 & Feb 2026 & & 0.4 \\
GPT-5.4~\citep{openai2026gpt54} & Aug 2025 & Mar 2026 & & 0.2 \\
Grok 4.20~\citep{xai2026grok} & Sep 2025 & Mar 2026 & & 0.1 \\
Claude Opus 4.7~\citep{anthropic2026opus47} & Jan 2026 & Apr 2026 & & 0.2 \\
GPT-5.5~\citep{openai2026gpt55} & Dec 2025 & Apr 2026 & & 0.4 \\
Claude Opus 4.8~\citep{anthropic2026opus48} & Jan 2026 & May 2026 & & 1.5 \\
Claude Fable 5~\citep{anthropic2026fable5} & Jan 2026 & Jun 2026 & 20$^\dagger$ & \\
Grok 4.5~\citep{xai2026grok} & Feb 2026 & Jul 2026 & 0.3 & \\
GPT-5.6 Luna~\citep{openai2026gpt56} & Feb 2026 & Jul 2026 & 0.0 & 0.2 \\
GPT-5.6 Terra~\citep{openai2026gpt56} & Feb 2026 & Jul 2026 & 2.3 & 0.8 \\
GPT-5.6 Sol~\citep{openai2026gpt56} & Feb 2026 & Jul 2026 & 13.3 & 7.8 \\
Grok 4.6~\citep{xai2026grok} & Feb 2026 & Aug 2026 & & 2.1 \\
\midrule
\rowcolor{humanrow} Human~\citep{arcprize2026arcagi3} & Mar 2026 & --- & 100 & 100 \\
\midrule
Claude Opus 5~\citep{anthropic2026opus5} & May 2026 & Jul 2026 & & 30.2 \\
GPT-6 Astra~\citep{openai2026gpt6} & Apr 2026 & Sep 2026 & & 62.7 (99.9)$^\S$ \\
\bottomrule
\end{tabular}
\end{table}

\subsection{DiG-bench}
\label{app:dig-background}

DiG-bench~\citep{battleday2026dig} comprises 70 text games across seven difficulty tiers, with one to sixteen levels per game.
Observations are strings of letters, numbers, and symbols, and actions are individual characters; both transition rules and win conditions must be inferred.
Limited lives and per-level step budgets make informative experiments essential: a game is won only when all levels are cleared before these budgets run out.
Many games also provide a creative mode for testing hypotheses outside the main level~\citep{digbench2026website}.
Three games per tier are public, while the remaining 49 are held privately for evaluation; our experiments cover all 21 public games.

\begin{table}[!ht]
\centering
\footnotesize
\setlength{\tabcolsep}{4pt}
\caption{Frontier models in the basic harness on all 70 DiG-bench games, as reported by the benchmark authors~\citep{battleday2026dig}. Win rates average runs within each game, then games within each set; tiers 6--7 contain 20 games. Each game was solved by at least one first-time human player.}
\label{tab:digbench}
\begin{tabular}{llccc}
\toprule
& & & \multicolumn{2}{c}{Win rate (\%)} \\
\cmidrule(lr){4-5}
Model & Cutoff & Release & All tiers & Tiers 6--7 \\
\midrule
Gemini 3.1 Pro~\citep{google2026gemini31pro} & Jan 2025 & Feb 2026 & 16.7 & 0.0 \\
Qwen 3.6 27B~\citep{qwen2026qwen36}          & Oct 2025 & Apr 2026 & 1.4  & 0.0 \\
GPT-5.5~\citep{openai2026gpt55}              & Dec 2025 & Apr 2026 & 25.7 & 10.0 \\
GLM-5.2~\citep{zai2026glm52}                 & Mar 2026 & Jun 2026 & 15.3 & 0.0 \\
Kimi K3~\citep{moonshot2026kimik3}           &          & Jul 2026 & 22.9 & 0.0 \\
Claude Opus 5~\citep{anthropic2026opus5}     & May 2026 & Jul 2026 & 71.4 & 40.0 \\
DeepSeek V4 Flash~\citep{deepseek2026v4}     &          & Jul 2026 & 11.4 & 0.0 \\
DeepSeek V4 Pro~\citep{deepseek2026v4}       & Apr 2026 & Aug 2026 & 4.3  & 0.0 \\
\midrule
\rowcolor{humanrow} Human solvability~\citep{battleday2026dig} & Aug 2026 & --- & 100 & 100 \\
\bottomrule
\end{tabular}
\end{table}

\subsection{MazeBench}
\label{app:maze-background}

MazeBench~\citep{pappas2026mazebench} is a three-dimensional exploration and puzzle environment with 256 interconnected rooms and 100 gems (Figure~\ref{fig:maze-env-appendix}).
Progress requires discovering mechanisms involving pushable blocks, lifts, bridges, and icy surfaces, then applying them across distant rooms.
We use the ASCII track, which renders the current room from a rotatable camera.
The interface exposes eleven actions for movement, camera control, undo, room reset, and return to visited rooms~\citep{pappas2026mazeintro}.
Gems collected and rooms visited measure puzzle-solving and exploration over long interactions, making this environment useful for studying how agents accumulate and reuse knowledge.

\begin{figure}[!ht]
\centering
\includegraphics[width=\linewidth]{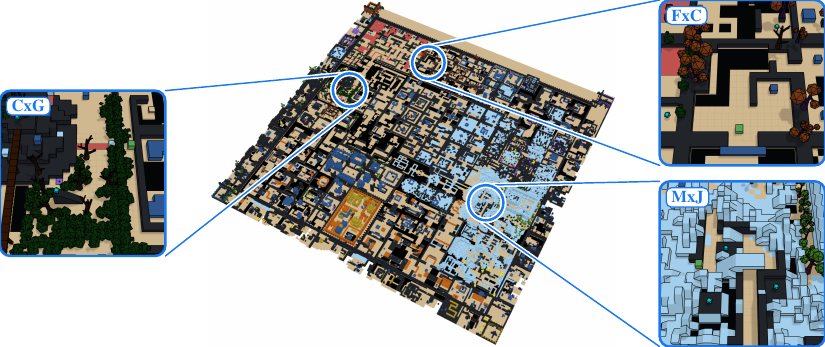}
\caption{\textbf{The MazeBench environment.} The full world of 256 interconnected rooms, with enlarged views of rooms CxG, FxC, and MxJ. Blue circles and connecting lines locate each room within the world.}
\label{fig:maze-env-appendix}
\end{figure}

\begin{table}[!ht]
\centering
\footnotesize
\setlength{\tabcolsep}{2.5pt}
\caption{Frontier models on MazeBench's ASCII track, from the September 14, 2026 leaderboard~\citep{pappas2026mazebench}. Tools permit Python execution. Gem percentages use each run's available total (71--100); rooms are out of 256. The human row gives the top-50 median, with gems out of 100; individual entries appear in Table~\ref{tab:maze-human-top50}.}
\label{tab:mazebench}
\begin{tabular}{llccccc}
\toprule
& & & \multicolumn{2}{c}{With tools} & \multicolumn{2}{c}{Without tools} \\
\cmidrule(lr){4-5} \cmidrule(lr){6-7}
Model & Cutoff & Release & Gems (\%) & Rooms & Gems (\%) & Rooms \\
\midrule
Gemini 3.1 Pro~\citep{google2026gemini31pro} & Jan 2025 & Feb 2026 & 1.1  & 5   & 0.0 & 1 \\
GPT-5.4~\citep{openai2026gpt54}              & Aug 2025 & Mar 2026 &      &     & 0.0 & 2 \\
GPT-5.5~\citep{openai2026gpt55}              & Dec 2025 & Apr 2026 &      &     & 0.0 & 4 \\
Claude Opus 4.8~\citep{anthropic2026opus48}  & Jan 2026 & May 2026 &      &     & 0.0 & 4 \\
GLM-5.2~\citep{zai2026glm52}                 & Mar 2026 & Jun 2026 &      &     & 0.0 & 1 \\
Claude Fable 5~\citep{anthropic2026fable5}   & Jan 2026 & Jun 2026 & 14.7 & 43  & 1.4 & 10 \\
GPT-5.6 Luna~\citep{openai2026gpt56}         & Feb 2026 & Jul 2026 & 0.0  & 2   & 0.0 & 1 \\
GPT-5.6 Terra~\citep{openai2026gpt56}        & Feb 2026 & Jul 2026 & 6.7  & 12  & 0.0 & 2 \\
GPT-5.6 Sol~\citep{openai2026gpt56}          & Feb 2026 & Jul 2026 & 17.1 & 67  & 1.4 & 4 \\
Kimi K3~\citep{moonshot2026kimik3}           &          & Jul 2026 & 1.3  & 4   & 0.0 & 2 \\
Claude Opus 5~\citep{anthropic2026opus5}     & May 2026 & Jul 2026 & 16.0 & 62  & 1.3 & 8 \\
DeepSeek V4 Pro~\citep{deepseek2026v4}       & Apr 2026 & Aug 2026 & 0.0  & 2   & 0.0 & 1 \\
Grok 4.6~\citep{xai2026grok}                 & Feb 2026 & Aug 2026 & 1.1  & 6   &     & \\
Claude Fable 5.1~\citep{anthropic2026fable51} & Jun 2026 & Sep 2026 & 11.1 & 33  & 2.0 & 12 \\
GPT-6 Astra~\citep{openai2026gpt6}           & Apr 2026 & Sep 2026 & 23.0 & 113 & 14.0 & 103 \\
\midrule
\rowcolor{humanrow} Human top-50~\citep{pappas2026mazebench} & Sep 2026 & --- & & & 29.0 & 137 \\
\bottomrule
\end{tabular}
\end{table}

\input{figures/maze_human_top50_table}

\clearpage
\endgroup
\raggedbottom
\section{Case Studies}
\label{app:case-study}

\subsection{ARC-AGI-3: Cases Behind the Ablations}
\label{app:case-arc-analysis}
The following cases expand the examples in Section~\ref{sec:arcagi3} using recorded interactions and program revisions from the Claude Opus 4.8 runs.
\input{figures/case_arc_ablation}

\clearpage
\subsection{DiG-bench: Discovering a Hidden String Rule}
\label{app:case-dig}
\input{figures/case_p3_discovery}

\clearpage
\subsection{MazeBench: Reusing a Rule After Long Interaction}
\label{app:case-maze}
\input{figures/case_maze_reuse}

\clearpage
\raggedbottom
\section{More Results}
\label{app:more-results}

\subsection{ARC-AGI-3 Per-Game Results}
\label{app:arc-pergame}

\begin{figure}[p]
\centering
\includegraphics[width=\linewidth]{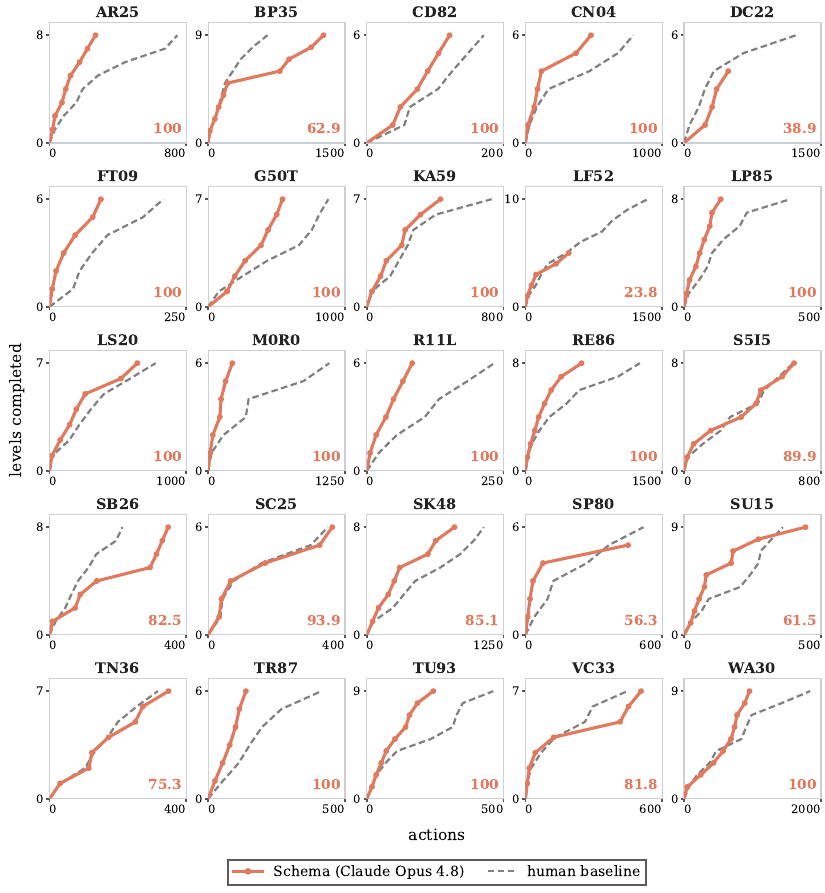}
\caption{Per-game progress on ARC-AGI-3 with Claude Opus 4.8 in Schema. Curves connect cumulative action counts at level completion for Schema and the human baseline. The number in each panel is the game's RHAE.}
\label{fig:arc-pergame-opus48}
\end{figure}

\begin{figure}[p]
\centering
\includegraphics[width=\linewidth]{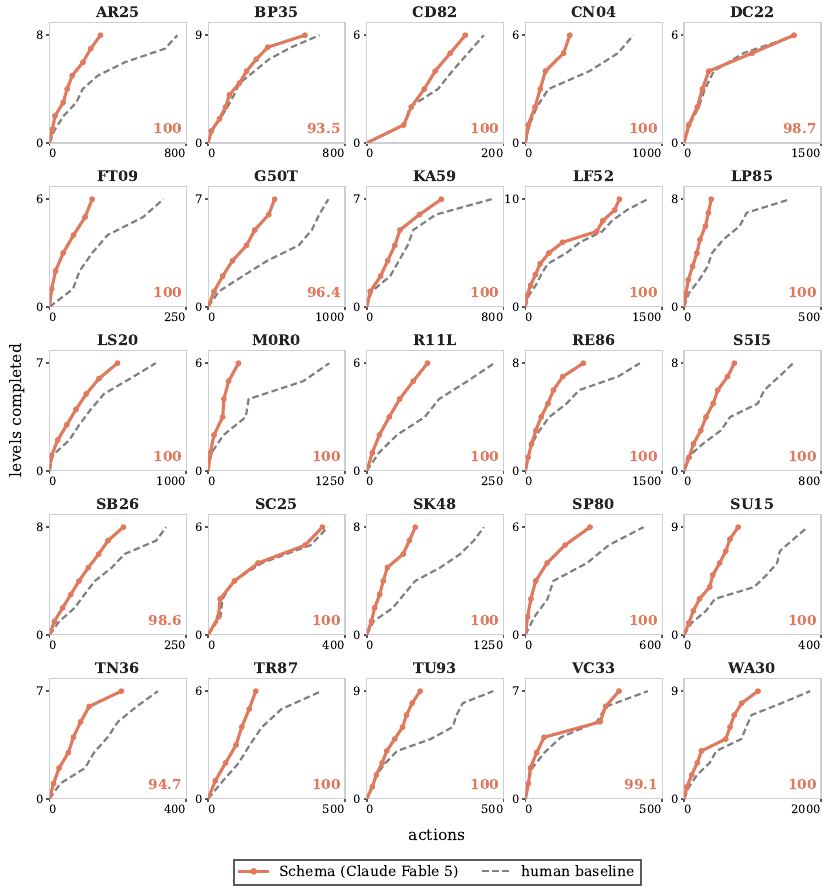}
\caption{Per-game progress on ARC-AGI-3 with Claude Fable 5 in Schema, as in Figure~\ref{fig:arc-pergame-opus48}.}
\label{fig:arc-pergame-fable5}
\end{figure}

\begin{figure}[p]
\centering
\includegraphics[width=\linewidth]{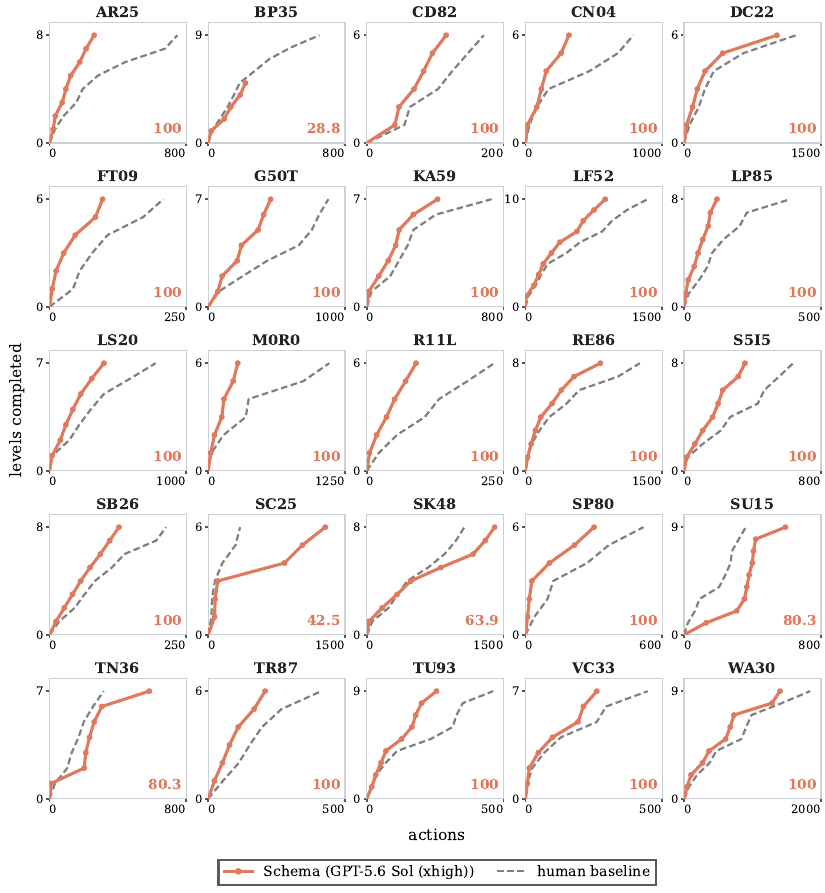}
\caption{Per-game progress on ARC-AGI-3 with GPT-5.6 Sol at extra-high effort in Schema, as in Figure~\ref{fig:arc-pergame-opus48}.}
\label{fig:arc-pergame-solxhigh}
\end{figure}

\begin{figure}[p]
\centering
\includegraphics[width=\linewidth]{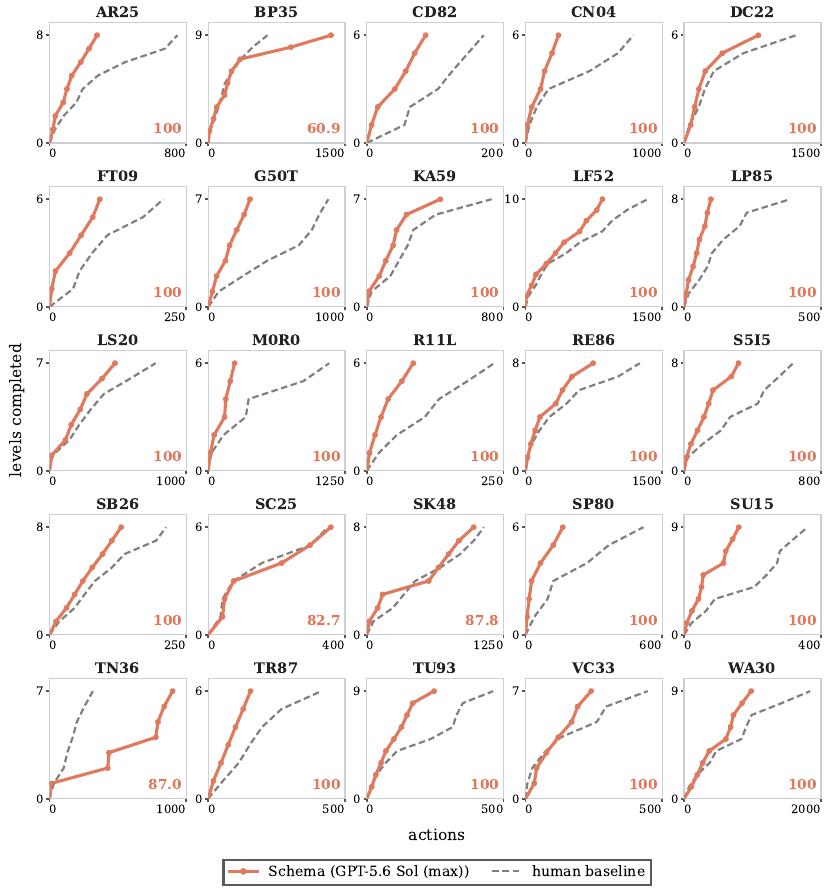}
\caption{Per-game progress on ARC-AGI-3 with GPT-5.6 Sol at maximum effort in Schema, as in Figure~\ref{fig:arc-pergame-opus48}.}
\label{fig:arc-pergame-solmax}
\end{figure}

Table~\ref{tab:arc-pergame} reports every public game for the four base models, and Figures~\ref{fig:arc-pergame-opus48}--\ref{fig:arc-pergame-solmax} trace each run level by level against the human baseline.
With Claude Fable 5, Schema clears all 183 levels in 57\% of the human baseline's actions and uses fewer actions than the human on 163 of them; with GPT-5.6 Sol at maximum effort it also clears every level, in 63\% of the human actions.

\input{figures/arc_pergame_table}

\clearpage
\subsection{DiG-bench Repeated Runs}
\label{app:dig-repeats}

\begin{figure}[!ht]
\centering
\includegraphics[width=\linewidth]{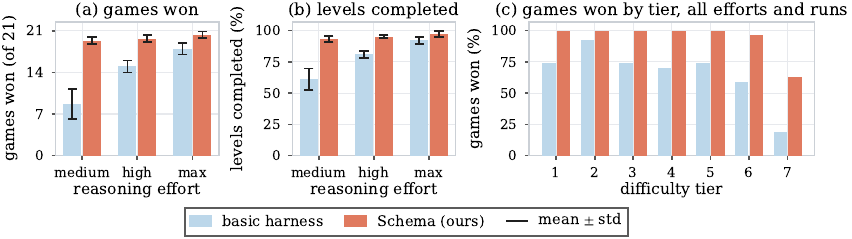}
\caption{DiG-bench with GPT-6 Astra over three runs. (a) Games won and (b) levels completed at each reasoning effort: bars show the mean and error bars one standard deviation. (c) Share of games won in each difficulty tier, over all efforts and runs.}
\label{fig:dig-repeats}
\end{figure}

\begin{figure}[!ht]
\centering
\includegraphics[width=\linewidth]{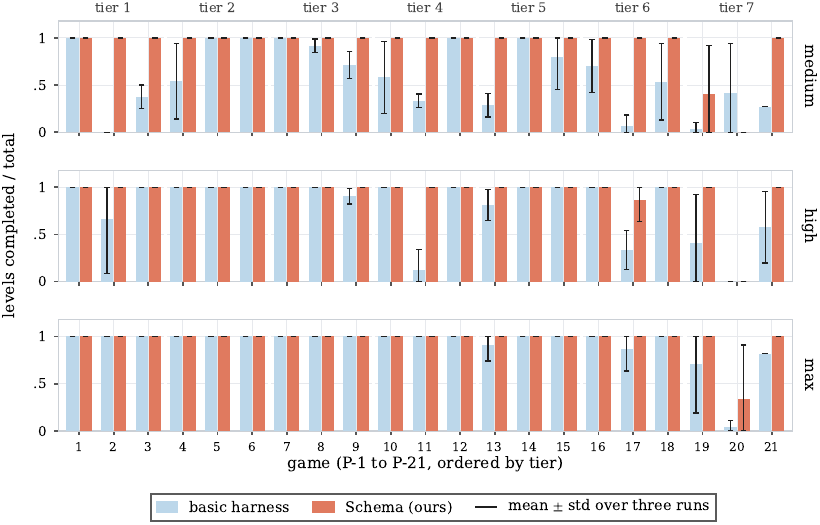}
\caption{Fraction of levels completed in each DiG-bench game at each reasoning effort, mean and standard deviation over three runs, games ordered by tier.}
\label{fig:dig-repeats-pergame}
\end{figure}

We ran every DiG-bench configuration three times with GPT-6 Astra.
Table~\ref{tab:dig-repeats} gives the mean and standard deviation of the games won and the levels completed, and Figures~\ref{fig:dig-repeats} and~\ref{fig:dig-repeats-pergame} break them down by reasoning effort, difficulty tier, and game.
Schema wins more games than the basic harness in every run at every reasoning effort, and its advantage is largest at medium effort, where it wins 19.3 games on average against 8.7; at medium effort it already matches the basic harness at maximum effort.
The gains come from the hardest games: across all efforts and runs, Schema wins 161 of the 162 game runs in tiers 1--6 and 17 of the 27 in tier 7, against 5 for the basic harness.

\input{figures/digbench_repeats_table}

\clearpage
\raggedbottom
\subsection{MazeBench Gem Collection and Room Coverage}
\label{app:maze-trajectory}

\begin{figure}[!ht]
\centering
\includegraphics[width=\linewidth]{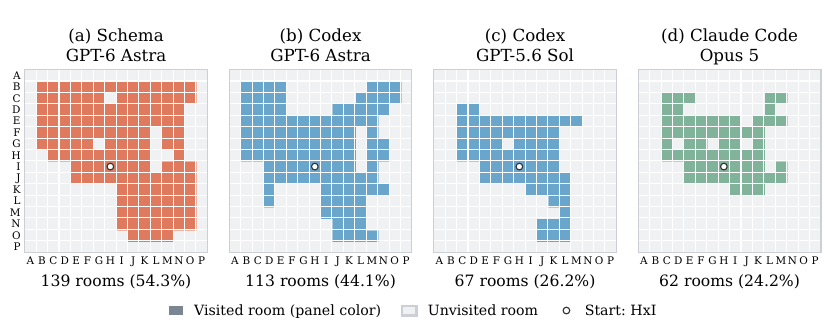}
\caption{\textbf{Room coverage of the four MazeBench runs.} Each cell is a room in the same $16\times16$ grid, with columns and rows A--P (e.g., CxG is column C, row G). Panels show each run's recorded endpoint. Baselines use the September 14, 2026 ASCII tools leaderboard, as in Figure~\ref{fig:maze}.}
\label{fig:maze-room-coverage}
\end{figure}

Figure~\ref{fig:maze-room-coverage} maps room coverage for the four trajectories in Figure~\ref{fig:maze}.
Schema and Codex with GPT-6 Astra share 106 visited rooms: 33 additional rooms appear only in Schema's trajectory and seven only in Codex's.
Schema also reaches every room visited by each of the other two baselines.

Table~\ref{tab:maze-gem-sequence} lists all 33 gems, collected across 31 rooms.
Coordinates are recovered from recorded top views and the gem's disappearance at collection.
Steps count all issued actions, including camera changes, undo, reset, and room returns.
Counts start at one: raw history index $i$ corresponds to step $i+1$, so the final index 27,818 represents 27,819 actions.

\input{figures/maze_gem_sequence_table}

\end{document}

%% file: math_commands.tex
\usepackage{amsmath,amsfonts,bm}

\def\eqref#1{equation~\ref{#1}}

\def\1{\bm{1}}

\DeclareMathAlphabet{\mathsfit}{\encodingdefault}{\sfdefault}{m}{sl}
\SetMathAlphabet{\mathsfit}{bold}{\encodingdefault}{\sfdefault}{bx}{n}

%% file: arxiv_style.tex
\arxiv{
\usepackage[a4paper, margin=2.5cm, headheight=28.5pt, headsep=3mm]{geometry}
\PassOptionsToPackage{hypertexnames=false}{hyperref}
\usepackage[colorlinks,hyperfootnotes=false]{hyperref}
\hypersetup{
    citecolor=[RGB]{50,100,170},
    linkcolor=[RGB]{50,100,170},
    urlcolor=[RGB]{255,102,178}}
\usepackage{fancyhdr}
}

\arxiv{
\newcommand{\toptitlebar}{%
  {\color{black}\hrule height 1pt}%
  \vskip 0.25in%
}
}

\makeatletter
\arxiv{
\renewcommand{\maketitle}{%
  \begin{center}%
    \toptitlebar
    \vskip 0.1in%
    {\LARGE\bfseries \@title \par}%
    \vskip 0.3in%
    {\normalsize \@author \par}%
  \end{center}%
  \par
  \vskip 0.3in%
}

\renewcommand\section{\@startsection {section}{1}{\z@}{-2.0ex plus
    -0.5ex minus -.2ex}{1.5ex plus 0.3ex minus .2ex}{\large\bfseries\raggedright}}
\renewcommand\subsection{\@startsection{subsection}{2}{\z@}{-1.8ex plus
    -0.5ex minus -.2ex}{0.8ex plus .2ex}{\normalsize\bfseries\raggedright}}
\renewcommand\subsubsection{\@startsection{subsubsection}{3}{\z@}{-1.5ex plus
   -0.5ex minus -.2ex}{0.5ex plus .2ex}{\normalsize\bfseries\raggedright}}

\renewenvironment{abstract}%
  {\centerline{\large\bfseries Abstract}%
   \begin{list}{}%
      {\setlength{\rightmargin}{0.6cm}%
       \setlength{\leftmargin}{0.6cm}}%
    \item[]\ignorespaces}%
  {\unskip\end{list}}

}
\makeatother

\iclr{
\PassOptionsToPackage{hypertexnames=false}{hyperref}
\usepackage[colorlinks=true,breaklinks=true]{hyperref}
}

\usepackage{hhline}

\makeatletter
\newcommand{\neutralize}[1]{\expandafter\let\csname c@#1\endcsname\count@}
\makeatother

\usepackage{algorithm}

\arxiv{
\usepackage{natbib}
\bibliographystyle{plainnat}
\bibpunct{(}{)}{;}{a}{,}{,}
}

\usepackage{amsmath}
\usepackage{mathtools}
\usepackage{amsthm}
\usepackage{amssymb}

\usepackage{xpatch}

\usepackage{thmtools}
\makeatletter
\@ifundefined{newcounteralias}{}{%
  \renewcommand\thmt@autorefsetup{%
    \expandafter\def\csname\thmt@envname autorefname\expandafter\endcsname
      \expandafter{\thmt@thmname}}}
\makeatother
\declaretheorem[name=Theorem]{theorem}
\declaretheorem[name=Lemma,sibling=theorem]{lemma}
\declaretheorem[name=Assumption,sibling=theorem]{assumption}
\declaretheorem[name=Condition,sibling=theorem]{condition}

\declaretheorem[name=Proposition,sibling=theorem]{proposition}

\makeatletter
  {%
   \pushQED{\qed}%
   \par\noindent{\bfseries\upshape {#1.}\ }%
  }%
  {\popQED\par}
  \makeatother

\theoremstyle{definition}

\theoremstyle{plain}
\newtheorem{corollary}[theorem]{Corollary}
\newtheorem{definition}{Definition}[section]

\usepackage[nameinlink,capitalize]{cleveref}

\crefformat{equation}{#2Eq.~(#1)#3}
\Crefformat{equation}{#2Eq.~(#1)#3}

\Crefformat{figure}{#2Figure~#1#3}
\Crefname{assumption}{Assumption}{Assumptions}
\Crefformat{assumption}{#2Assumption~#1#3}
\Crefname{subsubsection}{Section}{Sections}
\crefformat{subsubsection}{#2Section~#1#3}
\Crefformat{subsubsection}{#2Section~#1#3}
\crefname{algorithm}{Alg.}{Algs.}
\Crefname{algorithm}{Alg.}{Algs.}

\usepackage{crossreftools}
\usepackage{xparse}

\ExplSyntaxOn
\DeclareDocumentCommand{\XDeclarePairedDelimiter}{mm}
 {
  \__egreg_delimiter_clear_keys:
  \keys_set:nn { egreg/delimiters } { #2 }
  \use:x
   {
    \exp_not:n {\NewDocumentCommand{#1}{sO{}m} }
     {
      \exp_not:n { \IfBooleanTF{##1} }
       {
        \exp_not:N \egreg_paired_delimiter_expand:nnnn
         { \exp_not:V \l_egreg_delimiter_left_tl }
         { \exp_not:V \l_egreg_delimiter_right_tl }
         { \exp_not:n { ##3 } }
         { \exp_not:V \l_egreg_delimiter_subscript_tl }
       }
       {
        \exp_not:N \egreg_paired_delimiter_fixed:nnnnn 
         { \exp_not:n { ##2 } }
         { \exp_not:V \l_egreg_delimiter_left_tl }
         { \exp_not:V \l_egreg_delimiter_right_tl }
         { \exp_not:n { ##3 } }
         { \exp_not:V \l_egreg_delimiter_subscript_tl }
       }
     }
   }
 }

\keys_define:nn { egreg/delimiters }
 {
  left      .tl_set:N = \l_egreg_delimiter_left_tl,
  right     .tl_set:N = \l_egreg_delimiter_right_tl,
  subscript .tl_set:N = \l_egreg_delimiter_subscript_tl,
 }

\cs_new_protected:Npn \__egreg_delimiter_clear_keys:
 {
  \keys_set:nn { egreg/delimiters } { left=.,right=.,subscript={} }
 }

\cs_new_protected:Npn \egreg_paired_delimiter_expand:nnnn #1 #2 #3 #4
 {%
  \mathopen{}
  \mathclose\c_group_begin_token
   \left#1
   #3
   \group_insert_after:N \c_group_end_token
   \right#2
   \tl_if_empty:nF {#4} { \c_math_subscript_token {#4} }
 }
\cs_new_protected:Npn \egreg_paired_delimiter_fixed:nnnnn #1 #2 #3 #4 #5
 {
  \mathopen{#1#2}#4\mathclose{#1#3}
  \tl_if_empty:nF {#5} { \c_math_subscript_token {#5} }
 }
\ExplSyntaxOff

\XDeclarePairedDelimiter{\supnorm}{
  left=\lVert,
  right=\rVert,
  subscript=\infty
  }

%% file: figures/case_macros.tex
\definecolor{cardframe}{HTML}{A7B1BD}
\definecolor{cardtitle}{HTML}{EEF1F5}
\definecolor{diffadd}{HTML}{E3F4E7}
\definecolor{diffdel}{HTML}{FBE6E6}
\definecolor{diffaddfg}{HTML}{1E7B3A}
\definecolor{diffdelfg}{HTML}{B42318}
\definecolor{chipbg}{HTML}{F1F3F5}
\definecolor{termbg}{HTML}{0F1417}
\definecolor{termfg}{HTML}{E9ECEF}
\definecolor{termyes}{HTML}{8CE99A}
\definecolor{termno}{HTML}{FFA8A8}
\definecolor{okgreen}{HTML}{2F9E44}
\definecolor{misred}{HTML}{E03131}

\newtcolorbox{casecard}[4][4.25cm]{
  enhanced, sidebyside, sidebyside align=top seam, lefthand width=#1, sidebyside gap=3.2mm,
  colframe=cardframe, colback=white, boxrule=0.4pt, arc=1.2mm,
  segmentation style={draw=cardframe!70, line width=0.3pt, dashed},
  colbacktitle=cardtitle, coltitle=black, fonttitle=\small,
  title={\textbf{#2}\hspace{0.55em}{\color{black!55}#3}\hspace{0.9em}{\sffamily\bfseries #4}},
  boxsep=1.1mm, left=1.8mm, right=1.8mm, top=1.4mm, bottom=1.4mm, toptitle=0.5mm, bottomtitle=0.5mm,
  before skip=5pt, after skip=5pt, fontupper=\small, fontlower=\small
}

\newtcolorbox{caseskip}[1]{
  enhanced, colframe=cardframe!80, colback=cardtitle!55, boxrule=0.3pt, arc=1mm,
  borderline={0.3pt}{0pt}{cardframe!80, dashed}, frame hidden,
  boxsep=0.8mm, left=2mm, right=2mm, top=0.6mm, bottom=0.6mm, before skip=4pt, after skip=4pt,
  fontupper=\footnotesize, before upper={\textbf{#1}\hspace{0.7em}}
}

\newtcbox{\tchip}[1][chipbg]{on line, colback=#1, colframe=black!22, boxrule=0.25pt, arc=1.3pt,
  boxsep=0pt, left=2pt, right=2pt, top=1.1pt, bottom=1.1pt,
  fontupper=\fontencoding{T1}\selectfont\ttfamily\scriptsize}

  {\end{minipage}\par}

\newtcbox{\outbox}[1]{on line, tcbox width=auto limited, colback=#1!10!white, colframe=#1!55!white, boxrule=0.3pt, arc=1.3pt,
  boxsep=0pt, left=2.5pt, right=2.5pt, top=1.3pt, bottom=1.3pt, fontupper=\footnotesize}

\newtcolorbox{caseterm}{colback=termbg, colframe=termbg, arc=1mm, boxrule=0pt, boxsep=0pt,
  left=1.8mm, right=1mm, top=1.2mm, bottom=1.2mm, before skip=0pt, after skip=0pt,
  fontupper=\fontencoding{T1}\selectfont\ttfamily\scriptsize\color{termfg}}

%% file: figures/maze_human_top50_table.tex
\begin{table}[!ht]
\centering
\footnotesize
\setlength{\tabcolsep}{5pt}
\caption{Top-50 human entries used for the MazeBench reference, comprising submissions through September 19, 2026~\citep{pappas2026mazebench}. Names are omitted; gems and rooms are out of 100 and 256, respectively, and hours denote recorded playtime.}
\label{tab:maze-human-top50}
\begin{tabular}{rrrrr@{\hspace{2em}}rrrrr}
\toprule
Rank & Gems & Rooms & Actions & Hours & Rank & Gems & Rooms & Actions & Hours \\
\cmidrule(lr){1-5}\cmidrule(lr){6-10}
1 & 80 & 233 & 115,060 & 55.99 & 26 & 29 & 104 & 49,221 & 95.95 \\
2 & 80 & 233 & 151,663 & 27.21 & 27 & 26 & 139 & 40,112 & 12.31 \\
3 & 80 & 233 & 260,784 & 102.35 & 28 & 26 & 126 & 68,310 & 13.64 \\
4 & 79 & 233 & 63,543 & 10.82 & 29 & 21 & 174 & 27,181 & 22.36 \\
5 & 79 & 233 & 206,116 & 29.43 & 30 & 21 & 129 & 26,261 & 5.04 \\
6 & 79 & 233 & 221,343 & 72.30 & 31 & 21 & 105 & 31,332 & 11.62 \\
7 & 79 & 233 & 227,236 & 286.17 & 32 & 21 & 104 & 31,654 & 9.79 \\
8 & 77 & 233 & 169,758 & 99.46 & 33 & 20 & 72 & 21,845 & 23.48 \\
9 & 68 & 209 & 119,655 & 23.30 & 34 & 17 & 96 & 5,001 & 21.48 \\
10 & 60 & 210 & 160,616 & 77.97 & 35 & 16 & 70 & 25,278 & 2.80 \\
11 & 59 & 233 & 120,523 & 80.42 & 36 & 15 & 62 & 21,428 & 24.80 \\
12 & 55 & 199 & 141,337 & 76.70 & 37 & 14 & 96 & 30,209 & 5.78 \\
13 & 49 & 202 & 54,560 & 11.56 & 38 & 13 & 105 & 31,618 & 7.97 \\
14 & 49 & 173 & 47,275 & 51.49 & 39 & 13 & 42 & 9,311 & 4.76 \\
15 & 45 & 174 & 42,721 & 26.39 & 40 & 12 & 62 & 40,629 & 5.63 \\
16 & 45 & 155 & 62,784 & 225.78 & 41 & 10 & 33 & 11,231 & 2.93 \\
17 & 40 & 162 & 94,565 & 23.07 & 42 & 9 & 49 & 27,759 & 8.52 \\
18 & 38 & 171 & 66,138 & 15.01 & 43 & 9 & 47 & 8,299 & 9.74 \\
19 & 38 & 156 & 39,231 & 72.14 & 44 & 9 & 20 & 10,299 & 1.59 \\
20 & 37 & 153 & 34,971 & 9.78 & 45 & 8 & 104 & 23,064 & 10.40 \\
21 & 36 & 165 & 62,131 & 21.71 & 46 & 8 & 86 & 11,620 & 13.85 \\
22 & 33 & 135 & 51,825 & 15.06 & 47 & 8 & 65 & 4,734 & 4.94 \\
23 & 32 & 164 & 58,978 & 31.11 & 48 & 8 & 27 & 11,690 & 12.61 \\
24 & 30 & 209 & 43,640 & 61.86 & 49 & 7 & 57 & 38,219 & 9.04 \\
25 & 29 & 119 & 64,195 & 9.42 & 50 & 7 & 41 & 6,834 & 0.89 \\
\bottomrule
\end{tabular}
\end{table}

%% file: figures/case_arc_ablation.tex
\begin{figure}[h]
\centering
\includegraphics[width=\linewidth]{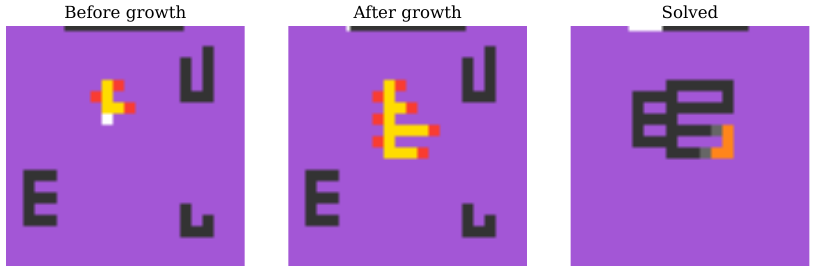}
\caption{Recorded CN04 frames before growth, after four growth actions, and at level completion. The final arrangement joins the pieces without overlapping their bodies.}
\label{fig:case-cn04}
\end{figure}

\paragraph{Program representation: assembling changing shapes (CN04, level 5).}
The agent moves and rotates pieces until their connectors meet, while keeping their bodies from overlapping.
One piece grows when acted on, changing both its outline and its connectors.
In the final search, Schema represents bodies and connectors as separate sets of grid cells and enumerates rotations and translations.
After four growth actions, the growing piece has seven connectors; the computed arrangement pairs all fourteen connectors into seven joints without any body overlap (Figure~\ref{fig:case-cn04}).
The run clears the level in 279 actions, compared with 1,219 for the prose-model variant.

\begin{figure}[h]
\centering
\includegraphics[width=\linewidth]{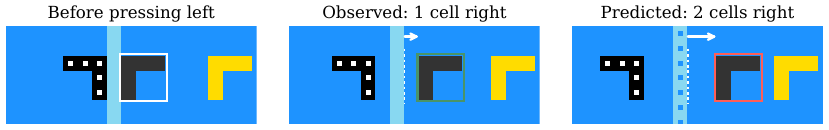}
\caption{The same historical AR25 transition, shown with identical crops. Boxes mark the reflection; dashed lines mark its original position. The turn-15 program predicts the observed one-cell movement exactly, while the turn-39 program predicts two cells.}
\label{fig:case-ar25}
\end{figure}

\paragraph{Certification: a new rule can break an old prediction (AR25).}
AR25 requires moving shapes and mirror axes so that the shapes and their reflections cover the targets.
To predict an arrow-key action, the program must identify which object is currently selected.
In the run without certification, a revision on level 3 addresses a piece whose selection marker is obscured by a target.
But the revised inference also changes predictions for the already completed first level: agreement on the same sixteen recorded transitions falls from 16/16 to 1/16.
Across the history available at the two checkpoints, agreement falls from 42/43 (97.7\%) to 30/46 (65.2\%).
Further revisions also introduce an error in the reflection geometry: moving a piece one cell left should move its reflection one cell right, but the later program moves the reflection two cells (Figure~\ref{fig:case-ar25}).
With certification, Schema's final model reproduces 267/268 transitions (99.6\%).

\begin{figure}[h]
\centering
\includegraphics[width=\linewidth]{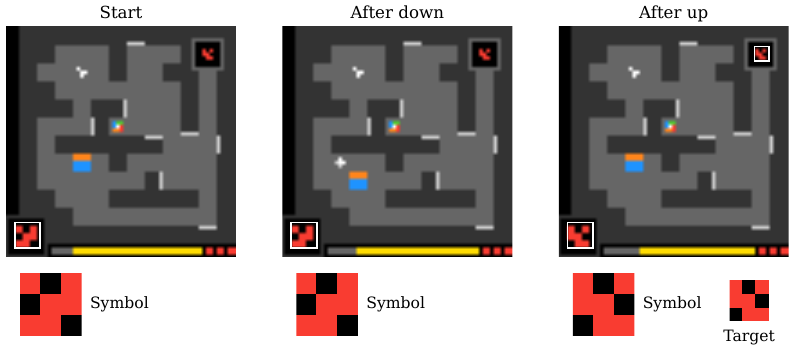}
\caption{Recorded LS20 observations before and after the first two actions of the final plan. Enlarged crops show the carried symbol rotating to match the target after the down--up pair. The remaining eleven actions reach the goal.}
\label{fig:case-ls20-plan}
\end{figure}

\clearpage
\paragraph{Planning: coordinating transformations and movement (LS20, level 5).}
The agent carries a symbol that must match the goal in both shape and color.
Contact with a moving marker rotates the symbol, so reaching the right location is insufficient: the player and marker must meet at the right time, with enough energy left to reach the goal.
The variant without planning enters this level with a program that passes all 252 historical transition checks, but works out its routes and transformations manually and takes 233 actions to finish.
Schema finishes in 65 actions.
Its final thirteen-action plan begins with a down--up pair that makes the player and moving marker converge, rotating the carried symbol into the target shape; the remaining eleven actions reach the goal (Figure~\ref{fig:case-ls20-plan}).

\begin{figure}[h]
\centering
\includegraphics[width=\linewidth]{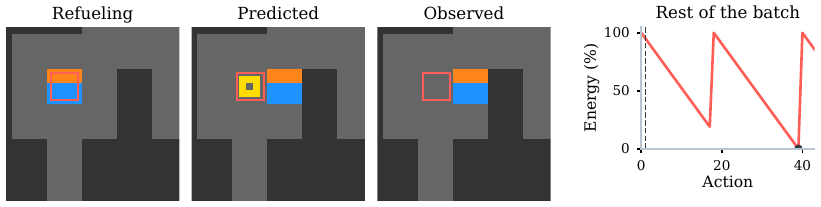}
\caption{The first prediction mismatch in the 43-action LS20 batch without verification. Identical crops highlight the station's location: the prediction retains it, but the observation shows floor. Execution continues through another refill and eventual energy depletion; the rebound after zero is an automatic reset.}
\label{fig:case-ls20-verify}
\end{figure}

\paragraph{Verified execution: noticing a spent resource (LS20, level 2).}
Here the agent must rotate its symbol and reach the goal while replenishing a limited energy supply.
The run without verification learns that yellow stations refill energy, but initially treats them as reusable.
It commits a 43-action plan that relies on returning to a previously used station.
The very first action already exposes the mistake: after the player leaves a station, the model predicts that the station remains, while the observation shows bare floor (Figure~\ref{fig:case-ls20-verify}).
Execution continues for all 42 remaining actions, including the return to the spent station, and the energy supply is exhausted before the batch ends.
Only afterward does the agent revise the program to consume a station after use and require a still-present station for refueling.

%% file: figures/case_p3_discovery.tex
\paragraph{The puzzle.}
P-3 asks the agent to discover which strings over \texttt{a}, \texttt{b}, \texttt{c}, and \texttt{d} satisfy a hidden rule.
Practice queries return yes or no without costing a life; the agent then takes an eight-question quiz, where each wrong classification costs a life.
We follow level 3 of a Schema run with GPT-6 Astra at medium effort.

\paragraph{From letter positions to letter counts.}
After \texttt{a} is accepted, the agent first encodes the hypothesis that accepted strings end in \texttt{a}.
The rejection of \texttt{ba} contradicts this rule, so it adds the requirement that the string also start with \texttt{a}.
The next queries expose the problem with that explanation: \texttt{aba} and \texttt{ac} are both accepted, although only the former ends in \texttt{a}.
The agent replaces the positional rule with \texttt{s.count('a') > s.count('b')}.
It then tests \texttt{baa}, \texttt{abba}, and \texttt{cad}: rearranging the letters preserves acceptance, adding a second \texttt{b} removes it, and surrounding \texttt{a} with other letters preserves it (Table~\ref{tab:p3-discovery}).

\begin{table}[h]
\centering
\small
\setlength{\tabcolsep}{5pt}
\renewcommand{\arraystretch}{1.25}
\begin{tabular}{@{}p{0.29\linewidth}p{0.30\linewidth}p{0.33\linewidth}@{}}
\toprule
Current hypothesis & New practice queries & Consequence \\
\midrule
Initial probe & \texttt{a}: yes & Propose a rule involving \texttt{a}. \\
Ends in \texttt{a} & \texttt{b}, \texttt{c}, \texttt{d}, \texttt{ab}, \texttt{ba}: no & \texttt{ba} rules out the current hypothesis. \\
Starts and ends in \texttt{a} & \texttt{aa}, \texttt{aba}, \texttt{ac}: yes & \texttt{ac} rules out this positional rule. \\
More \texttt{a}'s than \texttt{b}'s & \texttt{baa}: yes; \texttt{abba}: no; \texttt{cad}: yes & All three agree with the count rule. \\
\bottomrule
\end{tabular}
\caption{All twelve practice queries in P-3 level 3, grouped chronologically by the program hypothesis in use. Hypothesis descriptions summarize the recorded code edits.}
\label{tab:p3-discovery}
\end{table}

\paragraph{Checking the rule and taking the quiz.}
Each revision is checked against the accumulated interaction history.
After the final practice queries, the program reproduces all 107 checkable transitions collected so far, including the earlier levels and the string-entry mechanics.
Schema then classifies all eight quiz strings correctly without losing a life (Table~\ref{tab:p3-quiz}).
Across the full game, this run clears all eight levels with one life lost; the basic harness at medium effort stops at levels 3, 4, and 5 in our three runs.

\begin{table}[h]
\centering
\small
\setlength{\tabcolsep}{4pt}
\renewcommand{\arraystretch}{1.2}
\begin{tabular}{@{}lrrrrrrrr@{}}
\toprule
String & \texttt{dca} & \texttt{a} & \texttt{dbdacb} & \texttt{bdcaaa} & \texttt{dcbdba} & \texttt{acd} & \texttt{bcabca} & \texttt{adbcc} \\
\#\texttt{a} $-$ \#\texttt{b} & $1$ & $1$ & $-1$ & $2$ & $-1$ & $1$ & $0$ & $0$ \\
Answer & yes & yes & no & yes & no & yes & no & no \\
Correct & \checkmark & \checkmark & \checkmark & \checkmark & \checkmark & \checkmark & \checkmark & \checkmark \\
\bottomrule
\end{tabular}
\caption{The eight scored questions in order. The count difference is shown to make the learned rule easy to check.}
\label{tab:p3-quiz}
\end{table}

%% file: figures/case_maze_reuse.tex
\paragraph{The puzzle.}
On ice, one directional action can carry the player across several tiles.
The player stops when blocked or upon reaching ordinary ground, so solving an ice maze requires planning a sequence of stopping points rather than simply tracing a walkable path to the gem.
We follow the same Schema run with GPT-6 Astra from its first encounter with ice in room IxG to a later maze in room NxE.

\begin{figure}[h]
\centering
\includegraphics[width=\linewidth]{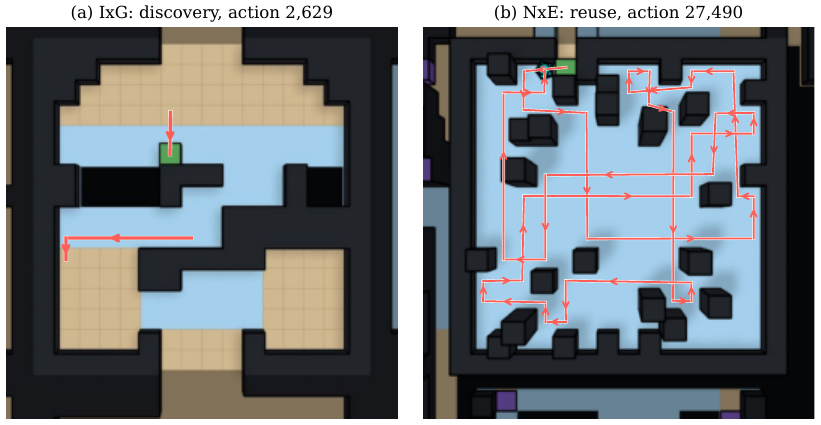}
\caption{Ice-rule discovery and reuse, shown on top-down game-engine views. Arrows trace selected probes in IxG (left), which stop at a wall or upon leaving ice, and the complete 38-action route to the gem in NxE (right).}
\label{fig:case-maze-ice}
\end{figure}

\paragraph{Learning a reusable movement rule.}
At action 2,629, a downward move in IxG slides the player across two tiles and stops just before a wall.
The agent adds sliding to its movement program and tests further directions.
Nine subsequent probes all match its predictions, including a slide that ends on the first ordinary tile beyond the ice (Figure~\ref{fig:case-maze-ice}a).
By action 2,638, the revised program reproduces all 2,638 recorded transitions.

\paragraph{Applying the rule in a new room.}
Schema first enters NxE at action 27,486.
From several viewpoints and an initial movement, it recovers the wall and ice geometry, including the ordinary entry tile initially hidden by the player.
Comparing the program immediately before entering NxE with the version used for the final plan shows that all 58 top-level functions and classes are unchanged: the additions import and register the new room's geometry.
The existing movement code supplies the consequences of sliding through this new layout.

\paragraph{Planning after repeated context compaction.}
At action 27,490, A* search finds a 38-action route to the gem, 24,852 actions after the IxG probes (Figure~\ref{fig:case-maze-ice}b).
The saved session records at least 35 context compactions between discovery and this plan.
All 38 actions execute, and each complete observation matches its saved forward prediction; the last action collects the gem at action 27,528.

%% file: figures/arc_pergame_table.tex
\begin{table}[!ht]
\centering
\small
\setlength{\tabcolsep}{3.8pt}
\caption{Per-game ARC-AGI-3 results of Schema with each base model. \emph{Levels} is the number of levels in the game and \emph{Human} the total actions of the human baseline; for each model, \emph{RHAE} is the game's score and \emph{Actions} counts actions through the last cleared level. Parentheses give completion counts for partially cleared games. $^\ast$The ten games with the most human actions, used for the ablations in Section~\ref{sec:arcagi3}.}
\label{tab:arc-pergame}
\begin{tabular}{@{}lrrrrrrrrrr@{}}
\toprule
 & & & \multicolumn{2}{c}{Opus 4.8} & \multicolumn{2}{c}{Fable 5} & \multicolumn{2}{c}{Sol xhigh} & \multicolumn{2}{c}{Sol max} \\
\cmidrule(lr){4-5}\cmidrule(lr){6-7}\cmidrule(lr){8-9}\cmidrule(lr){10-11}
Game & Levels & Human & RHAE & Actions & RHAE & Actions & RHAE & Actions & RHAE & Actions \\
\midrule
AR25$^\ast$ & 8 & 748 & 100.0 & 269 & 100.0 & 298 & 100.0 & 261 & 100.0 & 278 \\
BP35 & 9 & 651 & 62.9 & 1{,}265 & 93.5 & 566 & 28.8 & 218 {\scriptsize(5/9)} & 60.9 & 1{,}347 \\
CD82 & 6 & 171 & 100.0 & 121 & 100.0 & 144 & 100.0 & 116 & 100.0 & 86 \\
CN04$^\ast$ & 6 & 789 & 100.0 & 479 & 100.0 & 324 & 100.0 & 318 & 100.0 & 241 \\
DC22$^\ast$ & 6 & 1{,}228 & 38.9 & 485 {\scriptsize(4/6)} & 98.7 & 1{,}205 & 100.0 & 1{,}018 & 100.0 & 814 \\
FT09 & 6 & 208 & 100.0 & 94 & 100.0 & 78 & 100.0 & 97 & 100.0 & 92 \\
G50T$^\ast$ & 7 & 879 & 100.0 & 544 & 96.4 & 486 & 100.0 & 457 & 100.0 & 306 \\
KA59 & 7 & 730 & 100.0 & 431 & 100.0 & 436 & 100.0 & 414 & 100.0 & 430 \\
LF52$^\ast$ & 10 & 1{,}339 & 23.8 & 475 {\scriptsize(5/10)} & 100.0 & 1{,}030 & 100.0 & 875 & 100.0 & 844 \\
LP85 & 8 & 388 & 100.0 & 134 & 100.0 & 99 & 100.0 & 120 & 100.0 & 98 \\
LS20$^\ast$ & 7 & 776 & 100.0 & 642 & 100.0 & 497 & 100.0 & 398 & 100.0 & 479 \\
M0R0$^\ast$ & 6 & 1{,}107 & 100.0 & 221 & 100.0 & 278 & 100.0 & 271 & 100.0 & 242 \\
R11L & 6 & 233 & 100.0 & 83 & 100.0 & 111 & 100.0 & 90 & 100.0 & 85 \\
RE86$^\ast$ & 8 & 1{,}255 & 100.0 & 615 & 100.0 & 635 & 100.0 & 822 & 100.0 & 742 \\
S5I5 & 8 & 638 & 89.9 & 643 & 100.0 & 294 & 100.0 & 356 & 100.0 & 318 \\
SB26 & 8 & 213 & 82.5 & 347 & 98.6 & 135 & 100.0 & 127 & 100.0 & 131 \\
SC25 & 6 & 350 & 93.9 & 363 & 100.0 & 334 & 42.5 & 1{,}285 & 82.7 & 359 \\
SK48$^\ast$ & 8 & 1{,}070 & 85.1 & 801 & 100.0 & 443 & 63.9 & 1{,}402 & 87.8 & 977 \\
SP80 & 6 & 518 & 56.3 & 450 {\scriptsize(5/6)} & 100.0 & 283 & 100.0 & 301 & 100.0 & 164 \\
SU15 & 9 & 361 & 61.5 & 444 & 100.0 & 158 & 80.3 & 593 & 100.0 & 160 \\
TN36 & 7 & 317 & 75.3 & 348 & 94.7 & 210 & 80.3 & 584 & 87.0 & 900 \\
TR87 & 6 & 414 & 100.0 & 138 & 100.0 & 174 & 100.0 & 208 & 100.0 & 155 \\
TU93 & 9 & 462 & 100.0 & 243 & 100.0 & 195 & 100.0 & 255 & 100.0 & 246 \\
VC33 & 7 & 447 & 81.8 & 507 & 99.1 & 342 & 100.0 & 261 & 100.0 & 240 \\
WA30$^\ast$ & 9 & 1{,}843 & 100.0 & 956 & 100.0 & 1{,}080 & 100.0 & 1{,}403 & 100.0 & 980 \\
\midrule
Mean / total & 183 & 17{,}135 & 86.1 & 11{,}098 & 99.2 & 9{,}835 & 91.8 & 12{,}250 & 96.7 & 10{,}714 \\
Games won &  &  & 22/25 &  & 25/25 &  & 24/25 &  & 25/25 &  \\
\bottomrule
\end{tabular}
\end{table}

%% file: figures/digbench_repeats_table.tex
\begin{table}[!ht]
\centering
\small
\caption{DiG-bench with GPT-6 Astra over three runs: mean $\pm$ standard deviation of the games won (of 21) and the levels completed (of 193).}
\label{tab:dig-repeats}
\begin{tabular}{@{}llccc@{}}
\toprule
 & & Medium & High & Max \\
\midrule
Games won & Basic harness & 8.7 $\pm$ 2.5 & 15.0 $\pm$ 1.0 & 18.0 $\pm$ 1.0 \\
 & \cellcolor{schemarow}Schema & \cellcolor{schemarow}\textbf{19.3 $\pm$ 0.6} & \cellcolor{schemarow}\textbf{19.7 $\pm$ 0.6} & \cellcolor{schemarow}\textbf{20.3 $\pm$ 0.6} \\
\addlinespace[2pt]
Levels completed & Basic harness & 118.0 $\pm$ 16.7 & 156.3 $\pm$ 5.5 & 178.0 $\pm$ 5.2 \\
 & \cellcolor{schemarow}Schema & \cellcolor{schemarow}\textbf{179.7 $\pm$ 4.6} & \cellcolor{schemarow}\textbf{183.7 $\pm$ 2.3} & \cellcolor{schemarow}\textbf{187.7 $\pm$ 4.6} \\
\bottomrule
\end{tabular}
\end{table}

%% file: figures/maze_gem_sequence_table.tex
\begin{table}[!ht]
\centering
\caption{\textbf{All 33 gems collected by Schema.} Read down the left block, then the right. Room-local $(x,y)$ coordinates range from 0 to 15 in the unrotated top view, increasing rightward and downward. Step is the cumulative action count; $\Delta$ counts actions since the previous gem (from the start for gem 1), including exploration elsewhere.}
\label{tab:maze-gem-sequence}
\footnotesize
\setlength{\tabcolsep}{3pt}
\renewcommand{\arraystretch}{1.1}
\begin{tabular*}{\linewidth}{@{\extracolsep{\fill}}rlcrr@{\hspace{15pt}}rlcrr@{}}
\toprule
Gem & Room & $(x,y)$ & Step & $\Delta$ & Gem & Room & $(x,y)$ & Step & $\Delta$ \\
\midrule
1 & HxH & $(1,3)$ & 100 & 100 & 18 & ExF & $(2,6)$ & 11,173 & 730 \\
2 & GxH & $(10,1)$ & 303 & 203 & 19 & ExC & $(13,12)$ & 11,478 & 305 \\
3 & HxF & $(1,1)$ & 512 & 209 & 20 & DxG & $(14,2)$ & 12,469 & 991 \\
4 & GxF & $(2,9)$ & 749 & 237 & 21 & IxJ & $(7,14)$ & 12,643 & 174 \\
5 & FxE & $(11,2)$ & 1,001 & 252 & 22 & JxL & $(14,14)$ & 13,759 & 1,116 \\
6 & FxF & $(10,2)$ & 1,116 & 115 & 23 & IxE & $(1,10)$ & 15,700 & 1,941 \\
7 & IxI & $(1,13)$ & 2,731 & 1,615 & 24 & MxN & $(1,15)$ & 17,168 & 1,468 \\
8 & JxG & $(12,8)$ & 3,170 & 439 & 25 & NxN & $(8,8)$ & 17,815 & 647 \\
9 & JxE & $(1,1)$ & 3,463 & 293 & 26 & MxJ & $(3,13)$ & 18,779 & 964 \\
10 & MxD & $(12,5)$ & 4,163 & 700 & 27 & MxJ & $(12,11)$ & 19,535 & 756 \\
11 & JxH & $(4,5)$ & 4,425 & 262 & 28 & MxJ & $(5,5)$ & 19,554 & 19 \\
12 & GxI & $(4,12)$ & 4,744 & 319 & 29 & MxI & $(11,5)$ & 19,742 & 188 \\
13 & CxE & $(2,3)$ & 5,019 & 275 & 30 & OxK & $(4,11)$ & 20,247 & 505 \\
14 & DxH & $(6,5)$ & 5,182 & 163 & 31 & MxL & $(14,8)$ & 21,732 & 1,485 \\
15 & CxD & $(3,12)$ & 6,042 & 860 & 32 & FxB & $(13,13)$ & 26,497 & 4,765 \\
16 & FxC & $(0,4)$ & 7,244 & 1,202 & 33 & NxE & $(11,14)$ & 27,528 & 1,031 \\
17 & NxF & $(13,11)$ & 10,443 & 3,199 &  & & & &  \\
\bottomrule
\end{tabular*}
\end{table}